\documentclass[default]{sn-jnl}
\usepackage{geometry}
\usepackage{graphicx}
\usepackage{multirow}
\usepackage{multicol}
\usepackage{amsmath,amssymb,amsfonts}
\usepackage{mathrsfs}
\usepackage[title]{appendix}
\usepackage{textcomp}
\usepackage{manyfoot}
\usepackage{booktabs}
\usepackage{algorithm}
\usepackage{algorithmicx}
\usepackage{algpseudocode}
\usepackage{listings}
\usepackage{inputenc}
\usepackage{hyperref}
\usepackage{longtable}
\usepackage[section]{placeins}
\usepackage[justification=centering]{caption}
\usepackage{bbding}
\usepackage{pifont}
\usepackage{wasysym}

\usepackage[dvipsnames]{xcolor}

\newtheorem{definition}{Definition}
\usepackage{float}
\begin{document}
\title[The Paradox of Noise: An Empirical Study of Noise-Infusion Mechanisms to Improve Generalization, Stability, and Privacy in Federated Learning]{The Paradox of Noise: An Empirical Study of Noise-Infusion Mechanisms to Improve Generalization, Stability, and Privacy in Federated Learning}

\author*[1]{\fnm{Elaheh} \sur{Jafarigol}}\email{elaheh.jafarigol@ou.edu}
\author[2]{\fnm{Theodore B.} \sur{Trafalis}}

\affil*[1]{\orgdiv{Data Science and Analytics Institute}, \orgname{University of Oklahoma}, \orgaddress{\street{202 W. Boyd St., Room 409}, \city{Norman}, \postcode{73019}, \state{Ok}, \country{USA}}}
\affil[2]{\orgdiv{Industrial and Systems Engineering}, \orgname{University of Oklahoma}, \orgaddress{\street{202 W. Boyd St., Room 104}, \city{Norman}, \postcode{73019}, \state{OK}, \country{USA}}}

\abstract{In a data-centric era, concerns regarding privacy and ethical data handling grow as machine learning relies more on personal information. This empirical study investigates the privacy, generalization, and stability of deep learning models in the presence of additive noise in federated learning frameworks. Our main objective is to provide strategies to measure the generalization, stability, and privacy-preserving capabilities of these models and further improve them.
To this end, five noise infusion mechanisms at varying noise levels within centralized and federated learning settings are explored. As model complexity is a key component of the generalization and stability of deep learning models during training and evaluation, a comparative analysis of three Convolutional Neural Network (CNN) architectures is provided.
The paper introduces Signal-to-Noise Ratio (SNR) as a quantitative measure of the trade-off between privacy and training accuracy of noise-infused models, aiming to find the noise level that yields optimal privacy and accuracy. Moreover, the Price of Stability and Price of Anarchy are defined in the context of privacy-preserving deep learning, contributing to the systematic investigation of the noise infusion strategies to enhance privacy without compromising performance. Our research sheds light on the delicate balance between these critical factors, fostering a deeper understanding of the implications of noise-based regularization in machine learning. 
By leveraging noise as a tool for regularization and privacy enhancement, we aim to contribute to the development of robust, privacy-aware algorithms, ensuring that AI-driven solutions prioritize both utility and privacy.}

\maketitle

\section{Introduction}\label{intro}  
In a world dominated by data-driven decision-making, artificial intelligence has offered remarkable capabilities in a wide range of applications, from healthcare to finance, smart cities, and beyond. Machine learning models, particularly deep neural networks, are built on abundant personal data, such as health records, financial data, browsing history, etc., collected by governmental organizations and the private sector. Despite the growing popularity of deep learning across domains, there are still concerns related to the algorithms' ability to generalize, maintain stability, and ensure privacy protection against adversaries. As the new applications of artificial intelligence enter different aspects of our lives, the recognition of privacy as a fundamental human right has increased. This calls for the development of ethical and responsible learning frameworks. Without proper mechanisms, individuals are exposed to potential misuse of personal data and harm. Adhering to privacy protection policies, machine learning practitioners strive to develop tools that enable the use of sensitive data while maintaining privacy. If privacy concerns are addressed, organizations and practitioners can leverage sensitive data responsibly to harness the power of machine learning without exposing individuals to risks. Differential privacy is designed to provide strong privacy guarantees for data analysis. By adding noise to the data, the differential privacy guarantee ensures that an attacker cannot infer sensitive information from the released data. 

Despite its promising implications for ensuring data privacy, adding noise to the data can result in a loss of accuracy. Therefore, more complex models are utilized to address the decline in performance since they are better at distinguishing helpful information from the noise in the data. Increasing the number of layers and hidden units in the network results in more complex models and improved generalization. However, overly complex models run the risk of overfitting and performing poorly on unseen data. Moreover, such models are more sensitive to variations in the data and model, resulting in significant fluctuations in the output. 

While excessive noise can be disruptive, introducing controlled perturbations during training can contribute to improved privacy protection through techniques like differential privacy, generalization, and stability. 

The objective of this study is to evaluate this claim and develop a systematic method of fine-tuning the noise parameters to achieve the desired privacy protection guarantees without sacrificing the accuracy of the results. We focus on Convolutional Neural Networks (CNN) for image classification and delve into the challenges and strategies of noise infusion mechanisms in centralized and federated settings. The scope of this research study is outlined. The two research questions with their corresponding tasks have been proposed as follows: 
\newline
\textbf{Research Question 1:} How does the incorporation of noise in different locations within the model structure or the data affect training outcomes? 
\newline
Task 1: Comparison of three CNN architectures to assess the impact of model capacity on generalization and stability during training and evaluation in noisy conditions. 
\newline
Task 2: Comparison of training models with Gaussian noise hidden layers against other noise infusion mechanisms.
\newline
Task 3: Comparative analysis of training CNN models with Gaussian noise hidden layers under various noise levels in centralized and federated learning.
\newline
\newline
\textbf{Research Question 2:} How can we estimate the level of additive noise prior to detecting a significant model performance decrease?
\newline
Task 1: Introducing the Signal-to-Noise ratio to quantify the trade-off between increasing the noise level and training accuracy and to find the optimal balance between privacy and accuracy. 
\newline 
Task 2: Introducing the Price of Stability and Price of Anarchy to gain a measurable perspective on the trade-offs between model performance and privacy due to increasing noise levels.

Motivated by the potential benefits of noise, we explore the implications and limitations of training with noise to gain a deeper insight into the impact of noise on generalization, stability, privacy, and overall model performance. We combine structural stabilization and noise infusion mechanisms to improve the generalization and stability in deep neural networks while maintaining privacy. Proper architecture and regularization scheme balances the generalization power of the training model with its capacity to memorize the intricate patterns within the data without oversimplifying the model and possibly losing information. Enhanced by differential privacy, federated learning plays a pivotal role in the future of machine learning. As a collaborative framework, federated learning enables data processing without requiring the data to be centralized. Given the decentralized nature of data in federated learning, we can not utilize the sample size as we possibly could with aggregated data. Therefore, achieving stable models with great generalization is especially beneficial when working on unseen data distributed over multiple devices. Our findings shed light on the benefits of using noise to improve generalization, stability, and privacy. As federated learning provides a unique approach, the capacity of deep learning models to generalize beyond the training data while maintaining privacy and stability in the face of perturbations becomes more critical in real-world applications. By doing so, we hope to contribute to developing stable and differentially private algorithms, allowing them to generalize effectively and support federated learning.

The paper is organized as follows. Section \ref{sec-preliminary} discusses some background material related to generalization, stability, privacy, differential privacy, and federated learning. Section \ref{sec-noisey deep learning} explores the potential of training with noise in deep neural network architectures. We also delve into the description of the Signal-to-Noise ratio, Price of Stability, and Price of Anarchy and their applications. In section \ref{sec-result}, the outcome of the numerical experimentation with a discussion of the results is provided. The numerical analysis consists of four experiments in centralized and federated settings and multiple noise infusion mechanisms. Finally, section \ref{sec-conclusion} concludes the paper.
\section{Preliminaries}\label{sec-preliminary}
\subsection{Generalization}
Generalization is the model's ability to make accurate predictions about unseen data drawn from the same distribution as the training data. Generalization is measured by generalization error which is the difference between the training error and the test error. The generalization capability of the algorithms can be improved in three ways:
\begin{itemize}
\item Structural stabilization: This approach is based on adjusting the number of free parameters to control bias in the network. In deep learning tasks, structural stabilization is done by changing the number of hidden units or pruning the weights in the architecture.
\item Regularization: Controls the variance by applying modifications to the cost function and adding a penalty term.
\item Random noise injection: Empirical studies have shown that additive noise improves generalization in deep neural networks. Adding random noise behaves as a form of regularization, which prevents the model from getting too complex and memorizing the input data. Section \ref{sec-noisey deep learning} provides more details on this topic.   
\end{itemize}

In deep neural networks, generalization is impacted by the complexity and capacity of the model.
\subsubsection*{Rademacher Complexity}
Rademacher complexity \cite{bartlett2002rademacher, gnecco2008approximation} is a great tool for measuring the complexity of a learning algorithm.
Rademacher complexity is a quantitative way of measuring the complexity of a hypothesis class based on its ability to learn the random noise within the data and minimize the gap between the empirical risk and the true risk \cite{ledoux1991probability, mohri2008rademacher, mohri2018foundations}. 
\begin{definition}
Assuming that $S$ is a set of data  sampled from distribution $D$, with input $x_{i}$ and label $y_{i}$,  $S = ((x_{1},y_{1}),...(x_{m}, y_{m})) \sim D$, then the hypothesis class $H$ is the set of functions that maps input $x_{i}$ to $y_{i}$.
The empirical Rademacher complexity of $H$ over $S$ is defined as: 
\begin{equation}\label{rademacher-equation}
    \mathscr{R}_S(H) = E_{\sigma}\left[max_{h \in H}\frac{1}{N}\sum_{i = 1}^{N} \sigma_{i}h(x_{i})\right]
\end{equation}
where,
\begin{equation}
\sigma_{i} =
\begin{cases}
1 & \text{With probability 0.5}\\
-1 & \text{With probability 0.5}\\
\end{cases} 
\end{equation}
In this equation, $E_{\sigma}$ is the expectation over the Rademacher random variable $\sigma$.
\end{definition}

Rademacher random variable behaves similarly to a coin flip. Assuming that $S\prime \sim D$ is a ghost sample, the labels are flipped using the Rademacher random variable, which acts as introducing random noise into the data. The goal is to find a function that minimizes the gap between the true and empirical risks while classifying the new sample $S\prime$. Rademacher complexity evaluates the classifier's success in minimizing the gap between the empirical and true risks, denoted as $R(H) - \hat{R}(H)$. The idea behind Rademacher complexity is that maximizing the correlations between the output of the hypothesis and labels is equivalent to minimizing the training error in the presence of the Rademacher random variable. Empirical studies show that the correlation is more significant when the hypothesis space is more complex. 

Rademacher complexity measures the trade-off between the model's capacity to learn noise and generalizing to unseen data. Higher Rademacher complexity indicates that the classifier is better at memorizing the noise and more prone to overfitting. We can decrease model complexity by controlling the capacity to avoid this issue. 
\subsubsection*{Vapnik-Chervonenkis (VC) Dimension}
Model capacity, quantified by the VC dimension (Vapnik-Chervonenkis dimension) \cite{vapnik2013uniform}, is the network's ability to capture the underlying patterns and learn the intricate relationships within the data.
\begin{definition}

VC dimension of a set of functions is the largest set of finite data points that can be classified perfectly by the classifier. Hence, the training error of the model is zero. In other words, it is the maximum number of data points the classifier shatters in all possible ways.
\end{definition}

Classifiers with higher VC dimensions have higher capacity \cite{cybenko1996just, karpinski1995bounding}. 

Focusing on neural networks as learning algorithms, the model's capacity is correlated with the number and depth of fully connected layers and the interplay between the architecture and the non-linear activation functions \cite{sontag1998vc}.  Deep neural networks with multiple layers and millions of parameters have high capacity and VC dimension \cite{goodfellow2016deep}. 

\begin{figure*}[!ht]
\centering
\includegraphics[width = 1 \linewidth, keepaspectratio]{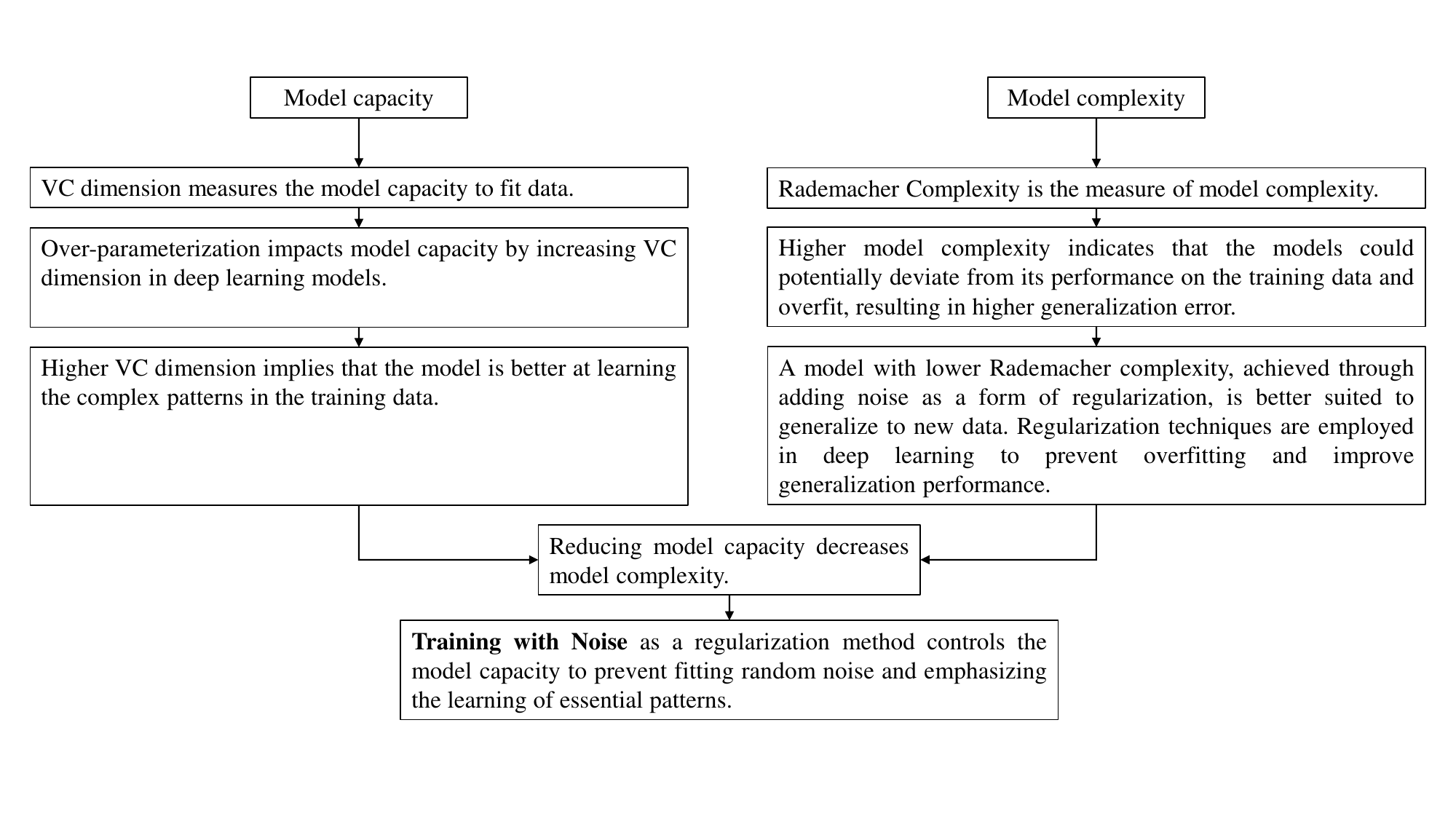}
\caption{The relationships between the VC dimension and Rademacher complexity allow for a more unified understanding of algorithm behaviors in nondeterministic circumstances in the presence of noise and the conditions leading to improved generalization.}
\label{fig-VC Rademacher}
\end{figure*}

High model capacity indicates that the model is capable of memorizing details from the training data and possibly overfitting when facing unseen data. Conversely, low model capacity results in an oversimplified model failing to fit the data properly. So, selecting the right architecture with sufficient model capacity is critical in deep learning. Figure \ref{fig-VC Rademacher} summarizes the interconnections between these concepts and how they influence each other in the context of deep learning and training with noise.

\subsection{Stability}
Stability is an essential property for learning algorithms. An algorithm is stable if the output of the algorithm doesn't change much when the training set is altered by one point, regardless of the sample size \cite{rakhlin2005stability} 
\begin{definition}
Lets assume $S = (z_{1}, z_{2},...,z_{i})$ and $S^\prime = (z_{1}, z_{2},...,z_{i},z^\prime$)  are two neighboring datasets that differ in one point $z^\prime$. For learning algorithm $h$, the loss function at point $z$ is denoted as $L_{z}(h)$. A learning algorithm is uniformly stable if equation \ref{eq-stability} holds for all $z$. 
\begin{equation}\label{eq-stability}
    \forall z \in z, |L_{z}(h_{s}) - L_{Z}(H_{S^\prime})| \le \beta 
\end{equation}
$\beta$ is the stability coefficient in this equation and is the smallest value that measures the difference between the loss of the algorithm on $S$ and $S^\prime$. Smaller $\beta$ corresponds to more stable algorithms. 
\end{definition}

Stability is closely related to the model's generalization ability on unseen data. Bousquet and Elisseeff \cite{bousquet2002stability} define the notion of stability for learning algorithms and demonstrate that stability is an algorithmic way of measuring generalization. Stable models are less prone to overfitting and have better generalization.
Stability is critical in designing practical learning algorithms, and a sensitivity analysis is the means to measure stability. This method, also known as perturbation analysis, is conducted by measuring the changes in the algorithm output in the presence of noise. Perturbation analysis allows us to utilize noise to design models capable of learning the underlying systems that produce data rather than the data itself \cite{bonnans1998optimization}. Sensitivity analysis is an essential component in defining generalization, stability, and differential privacy.
\subsection{Privacy}
In the modern world, where governments and private companies frequently use data for strategic planning, decision-making, policies, and even services, privacy is a serious concern. Privacy is the individual's autonomy in collecting, storing, sharing, and analysis of personal data. Privacy violations can have serious personal and social implications for vulnerable populations, causing discrimination, surveillance, and other potential harms. Emerging technologies in data generation, storage, and analysis raise new concerns about individuals' right to privacy in the machine learning domain. Motivated by the Fundamental Law on Information Reconstruction, the researchers in Microsoft Research Lab focused on designing a holistic approach to preserving privacy in the statistical learning of individuals' data. However, without a structured definition of privacy, evaluating the privacy-preserving methods was subject to failure. An intuitive definition of privacy is the one by Gavison\cite{gavison1980privacy}.
\begin{definition}\label{privacy-definition}
Privacy is the protection from being brought to the attention of others.
\end{definition}

As governments and organizations strive to harness the potential knowledge and value in the data, reliable and trustworthy algorithms become crucial. Researchers encourage policymakers to incorporate privacy as a human right in the processes and establish privacy protection mechanisms that ensure individuals' safety in the age of artificial intelligence \cite{tene2011privacy, brandeis1890right}. 
\subsection{Differential Privacy}
One of the most stringent measures of privacy is differential privacy, which ensures that adding or removing any individual's data does not change the probability of an outcome by "too much". 

The definition of differential privacy relies on the concept of a randomized algorithm, which has been employed in various applications, including cryptography and accelerating solutions of algebraic equations. Randomized algorithms are computational procedures that incorporate random choices or probabilistic decisions to solve problems. Rather than following a deterministic path, these algorithms leverage randomness, either to simplify the process or to achieve a solution with high probability. For example, a randomized algorithm can use a random event, such as flipping a coin as part of its description, and make decisions based on the outcome of the coin flips. Therefore a randomized algorithm maps inputs to probabilities of different outputs rather than deterministically mapping inputs to specific outputs.

A key benefit of differential privacy is providing mathematically rigorous privacy guarantees. Therefore, any particular algorithm's privacy protection level is clearly understood. The mathematical definition of privacy provides a measurable term for evaluating and maintaining privacy \cite{dinur2003revealing, blum2005practical, dwork2006calibrating, dwork2011differential}.
\begin{definition} \label{dp-definition}
Let's assume $x$ and $y$ are two datasets; the $l_{1}$ norm of  dataset x, denoted as $||x||_{1}$, is a measure of the size of the dataset, and it is defined to be:
\begin{equation}\label{L1-norm}
    ||x||_{1} = \sum_{i=1}^{|x|} |x|_{i},	
\end{equation}
The $l_{1}$ distance between $x$ and $y$ is  $||x - y||_{1}$, which measures the difference between the number of records in $x$ and $y$. Datasets are also perceived as a multiset of rows, so the distance between the datasets can be measured by the Hamming distance; that is, the difference in the number of rows between $x$ and $y$. \newline
$M$ is a randomized mechanism with domain  ${\mathbb{N}}^{|x|}$. \newline
$S$ is the set of outcomes of $M$; therefore, $S\in Range(M)$. \newline
Differential privacy is defined on two neighboring datasets. $x$ and $y$ are two neighboring datasets if the two datasets differ by only one sample (row). Hence, for all $x$ and $y$, the $l_{1}$ distance is $||x - y||_{1} \le 1$. $M$ is $\epsilon$-Differentially Private ($\epsilon$-DP) if equation \ref{dp-definition} holds for any two neighboring datasets derived from the dataset:
\begin{equation}\label{dp-equation}
    P[M(x) \in S] \le exp(\epsilon) P[M(y) \in S]
\end{equation}
\end{definition}

This definition is the strict definition of $\epsilon$-DP, and it has been studied explicitly in the book published on differential privacy by Dwork and Roth \cite{dwork2014algorithmic}. Differential privacy can be adjusted using a parameter that measures the desired privacy levels. In this definition, $\epsilon$ is a very small value known as privacy loss or leakage. $\epsilon$ determines the acceptable change in the output of the mechanism due to the inclusion or removal of any individual, so information learned about the individual as a result of participating in the dataset is limited. A relaxed version of this definition, currently used in most applications of differential privacy, is $(\epsilon, \delta)$-DP provided in Equation \ref{dp-relaxed-equation}. 
\begin{equation}\label{dp-relaxed-equation}
    P[M(x) \in S] \le exp(\epsilon) P[M(y) \in S] + \delta
\end{equation}
In this definition, $\delta$ is the probability of leaking more information than what $\epsilon$ claims. $\delta$ is preferably zero or a very small value, typically the inverse polynomial of the sample size denoted as $\delta = 1/n^k$ where $n$ is the sample size, and $k$ is a positive integer. This implies that a larger sample size reduces the risk of unintentional disclosure of private information resulting from a query. To achieve ($\epsilon$, $\delta$)-DP, additive noise is conditioned on the type of noise we are adding, the desired $\epsilon$ and $\delta$, the sample size, the number of queries performed on the database, and the desired accuracy. In differential privacy, computations involving noise safeguard personal data and prevent it from being reverse-engineered from the results \cite{wei2020federated}. However, leaking private information due to statistical queries and machine learning models compromises privacy \cite{dwork2006our}. Sensitivity is used to monitor this leakage of information.  
\begin{definition}
Sensitivity is the maximum change in the output of a query as a result of removing an individual from the database.
\end{definition}
Sensitivity is measured based on the distance ($d$) between the output of mechanism $M$ on the neighboring datasets $x$ and $y$, where $d(x, y) \le 1$. Sensitivity is defined as: 
\begin{equation}\label{eq-sensitivity}
\begin{split}
    & Sensitivity = max||M(x) - M(y)||_{1}
\end{split}
\end{equation}

Sensitivity helps characterize the impact of individual data on the output, while $\epsilon$ quantifies the upper bound on the level of privacy protection that the algorithm can guarantee.

In practice, differentially private algorithms are required to randomize the query or training model output by adding noise before publicly communicating it with other users. Under differential privacy, we must carefully choose where to add noise and select the appropriate type and amount. A common approach is adding noise sampled from a Gaussian distribution with a mean of $\mu = 0$ and a standard deviation of $\sigma$. A higher noise level provides stronger privacy guarantees. We can design private models that abide by the definition of differential privacy and are restricted under the desired privacy guarantees. In recent years, differential privacy has been widely used in the federated learning framework. 
\subsection{Federated Learning}
Federated learning is a promising paradigm for collaborative model training across multiple devices without data sharing  \cite{mcmahan2017communication}. Keeping the data decentralized reduces the risk of leakage and data breach \cite{bonawitz2019towards}. This ensures that the benefits of machine learning can be utilized without compromising the privacy of individuals or organizations. The training process starts by sending the global base model to a subset of data centers. The model is trained locally, and the parameters are securely transmitted to the global server. The parameters received from the data centers at each round of training are aggregated in the global server. The model is updated and sent to the data centers for training \cite{kairouz2021advances}.

Apart from keeping data decentralized, differential privacy is used to provide privacy by adding noise during training and sending perturbed parameters to the global server. Federated learning has a wide range of applications in healthcare, protection of genomics data, social sciences, finance, information collected on personal devices such as location, browsing history, user activities on the web, and many more. 
\subsection{Highlights}
Understanding the intricacies of machine learning models' ability to generalize is rooted in several key concepts. The main takeaways of this section for deep learning and privacy are provided. 
\newline 
1. Interplay of VC dimension, Rademacher complexity, stability, and generalization:  The notions of VC dimension, Rademacher complexity, and stability are closely intertwined and essential to the model's generalization ability. 
\begin{itemize}
    \item Rademacher complexity and stability encapsulate the algorithm's behavior towards noise in the data. While stability measures the changes in the model output in the presence of noise, Rademacher complexity quantifies the model's ability to learn the random noise in the data, and it is upper bounded by the VC dimension.
    \item Research by Ron and Kearns \cite{ron1999algorithmic} on the connection between VC dimension and stability indicates that for algorithms with finite VC dimensions, stability is bounded by the VC dimensions.
    \item Studies on the relationship between VC dimension and Rademacher complexity in deep neural network models by Neyshabour et al. \cite{neyshabur2017exploring} and Karpinski and  Macintyre \cite{karpinski1995bounding} suggest that VC dimension, Rademacher complexity, and the number of parameters are equivalent. Hence, the number of model parameters determines the model capacity.
    \item Deep learning models are said to be over-parameterized if the number of parameters is significantly larger than the number of available data points in the training set.
    \item Over-parameterized models are more prone to overfitting due to increased model capacity. 
    \item Large, diverse data can mitigate the risk of overfitting caused by over-parameterization. The abundance of data allows the model to learn the underlying patterns beyond the noise and perform well on unseen data. In situations with limited data, regularization techniques can be employed to prevent overfitting and enhance the generalization capability of a model. Regularization techniques control the variance by modifying the cost function and applying a penalty term. 
    \item Bishop \cite{bishop1995training} demonstrated that the regularization term is written as a Tikhonov regularizer in a simple neural network architecture with one input and one output. Tikhonov regularization is often referred to as ridge regression or $L_{2}$ regularization in machine learning. Bishop \cite{bishop1995neural} also
    \item Bishop \cite{bishop1995neural} also highlights that training with noise is a form of regularization in neural network models. His findings and the research by others, such as Shalev-Shwartz and Ben-David \cite{shalev2014understanding}, suggest that regularization results in stable algorithms.
    \end{itemize}
    
Careful regularization and architectural choices are essential to finding the balance between model complexity, stability, and generalization. Research Question 1 aims to explore this intricate balance further and provide insights on how to improve it. 
\newline
2. Stability and Differential Privacy:
\begin{itemize}
    \item Stability is a desirable property in machine learning models, as it ensures that minor changes in the input do not result in drastic changes in the output predictions.
    \item The definition of differential privacy inherently aligns with stability. Maximizing stability in algorithms offers stronger privacy protection guarantees under differential privacy. 
    \item A potential drawback of differential privacy is its negative impact on accuracy due to introducing noise during training. Excessive noise during training can disrupt the data and cause loss of information, leading to reduced model performance.
\end{itemize}

Careful tuning of the noise parameters is a critical step in training with noise. The optimal amount of noise can vary depending on factors such as the problem, the data, and the desired properties of the training model. Research Question 2 aims to provide solutions that can help improve the tuning process and enable the selection of an optimal amount of noise for a given problem and dataset.
\section{Training with Noise in Deep Neural Networks}\label{sec-noisey deep learning}
\begin{figure*}[!ht]
\centering
\includegraphics[width = 1 \linewidth, keepaspectratio]{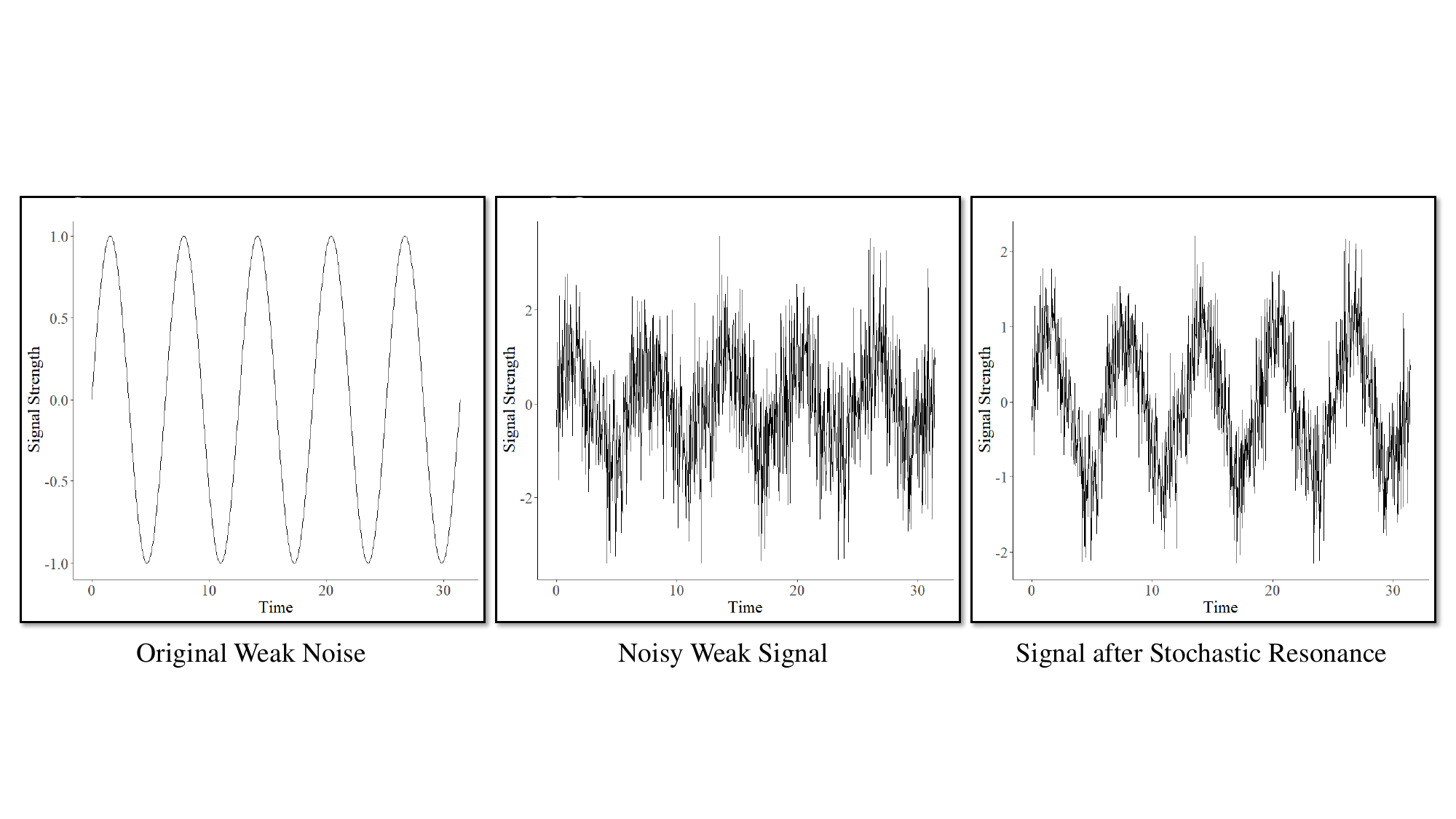}
\caption{The weak signal becomes more distinguishable after stochastic resonance. }
\label{Stochastic resonance}
\end{figure*}

Noise infusion has been studied in various domains. This phenomenon, known as \textit{stochastic resonance}, employs Gaussian noise to enhance the system's signal detection capabilities \cite{mcdonnell2011benefits, doyle2018colour, kumar2019novel}. The idea of stochastic resonance dates back to the early 1980s when Benzi et al. \cite{benzi1981mechanism, benzi1983theory}
introduced the phenomena and investigated its effect on complex systems. Figure \ref{Stochastic resonance} demonstrates the impact of Gaussian noise on amplifying the weak signals.

When the noise magnitude is small, additive noise enhances weak signals and improves the system's ability to identify useful data without negatively impacting the input. It also helps biological systems to adapt and learn from noisy environments \cite{ikemoto2018noise}. Stochastic resonance has a wide range of applications in science and engineering, from neuroscience to biological processes, signal processing, and information transmission. Numerous studies focus on the benefits of additive noise in pattern recognition in the nervous system and how it applies to computational neural network settings\cite{faisal2008noise,maass2014noise}.

\begin{figure*}[ht]
\centering
\includegraphics[width = 1 \linewidth, keepaspectratio]{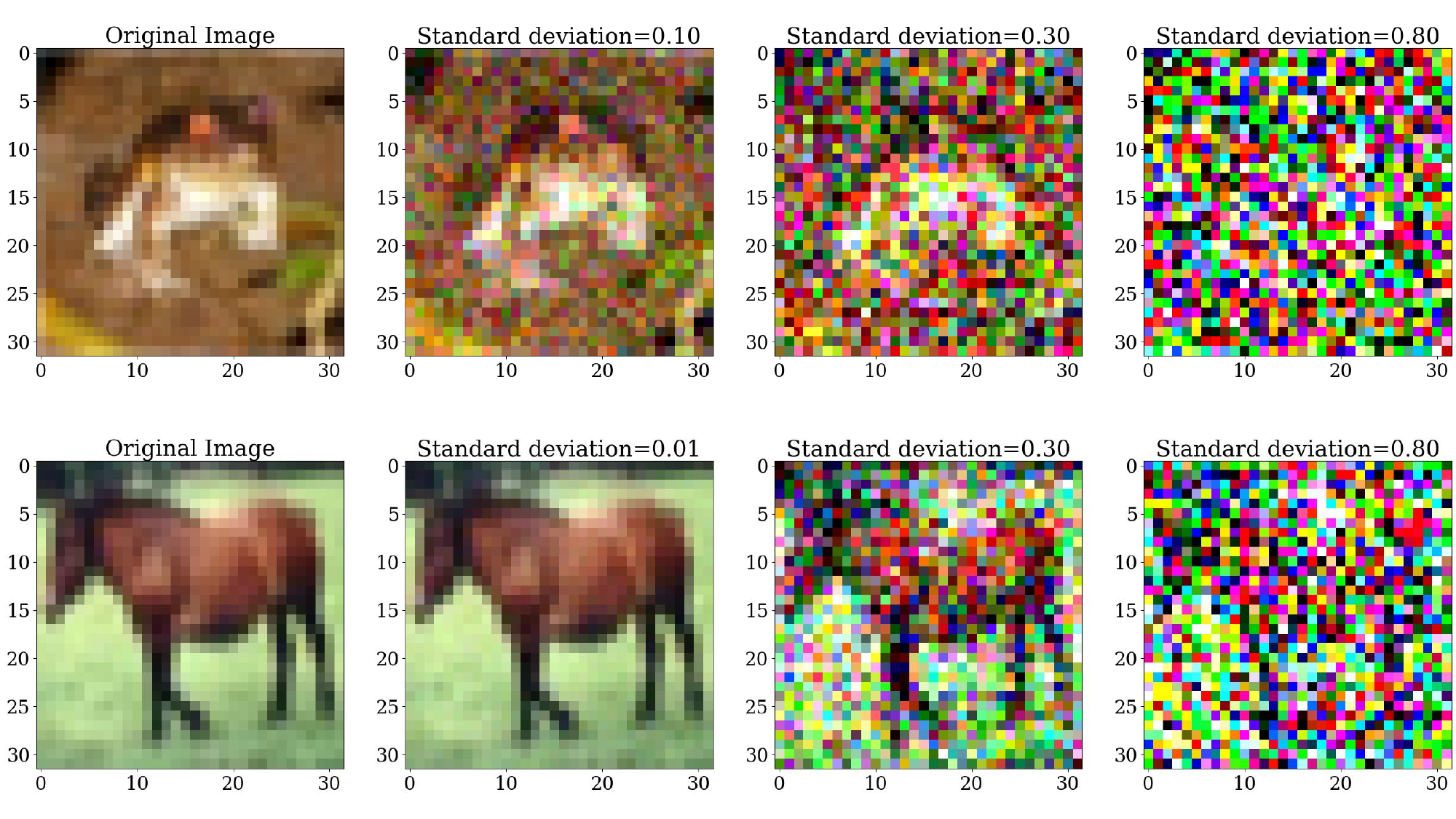}
\caption{Perturbed images with Gaussian noise.}
\label{Noise level}
\end{figure*}

Adding noise to a dataset alters the output of the queries. Figure \ref{Noise level} demonstrates the impact of input noise on two images taken from the CIFAR-10 dataset. The input noise is implemented by adding a random value sampled from the Gaussian distribution with a standard deviation of $\sigma$ during training. It can be observed that the images can absorb different noise magnitudes before they are completely corrupted. The problem specifications, data, and training models contribute to determining the appropriate noise level for training. 

Deep neural networks can learn the complex relationships in the data, making them well-suited for tasks such as image and speech recognition, natural language processing, and many other applications in artificial intelligence and machine learning. 

Despite their popularity, they are not a silver bullet that can solve all problems in artificial intelligence. Deep learning models are notoriously data-hungry and require a large amount of data to train on. Therefore, their performance relies on the intricacy of the problem and the data, model architecture, and optimization techniques. Their sensitivity to changes in the data distribution and complexity of the model architecture affects their ability to identify critical information rather than memorize the data. Failure in learning leads to overfitting the data. The benefits of adding noise during training are but are not limited to the following:
\begin{itemize}
\item Handling inherently noisy data as a result of measurement errors or corrupted data  \cite{holmstrom1992using, karpukhin2019training}
\item Handling inadequate training data for training: Noise infusion is an effective data augmentation method \cite{sietsma1991creating}. Noise infusion schemes help diversify the data collected on edge devices to improve the distributed learning results \cite{zeng2021noise}. 
\item Reducing overfitting and improving generalization: Empirical studies demonstrate that additive noise improves generalization in deep neural networks by preventing the model from getting too complex and memorizing the input data.  \cite{reed1999neural, zur2009noise, nagabushan2016effect, dhifallah2021inherent}. 
\item Improving the robustness of the neural network model against adversarial noise \cite{he2019parametric, xiao2021noise, liu2021training}. 
\end{itemize}

In deep learning, we can introduce noise into the algorithm by perturbing input, labels, gradients, weights, or the network's architecture. Table \ref{literature} presents some of the studies on noise infusion mechanisms in deep learning. The choice of the amount of noise and noise infusion mechanism is critical in designing an efficient model with the desired stability and generalization ability.
\begin{table}[ht]
\caption{Noise infusion mechanisms in deep learning literature} 
\centering
\begin{tabular}{|l|l|}
\hline
Noise Infusion Mechanisms & References \\ 
\hline
Input & \cite{koistinen1991, matsuoka1992noise, abadi2016deep, an1996effects, wang1999training, azamimi2010effect, alonso2014combining, isaev2016training, kosko2020noise, brown2003use, hua2006noise, li2016whiteout, lim2021noisy}\\
Hidden Layers & \cite{arani2021noise, you2019adversarial}\\
Model Weights & \cite{adilova2018introducing, sapkal2018modified, shi2020anti, bykov2021noisegrad}\\
Gradients & \cite{edwards1998fault, zhou2019toward, duan2021noise, chaudhari2015effect,neelakantan2015adding}\\
Labels& \cite{mirzasoleiman2020coresets, jiang2022towards, wu2022stgn, song2022learning}\\
\hline
\end{tabular}
\label{literature}
\end{table}

In this paper, we explore various noise infusion mechanisms for image classification using CNN. CNNs are a class of deep learning models designed primarily for processing and analyzing visual data, such as images and videos. They have revolutionized computer vision and have found widespread applications in various fields. CNN uses convolutional layers to automatically learn hierarchical features from input data, making them well-suited for tasks like image classification, object detection, and facial recognition. CNNs continue to evolve, pushing the boundaries of what is possible in machine perception and understanding. 
\subsection{Signal-to-Noise Ratio}
Signal-to-Noise Ratio (SNR) quantifies the clarity of the desired signal in the presence of noise in the signal processing domain. The idea of SNR is closely related to stochastic resonance, in which additive noise enhances weak signals\cite{smith1997scientist, mazda2014telecommunications}. SNR is defined as:
\begin{equation}\label{eq-snr}
    SNR = 10\times \log_{10}\frac{Signal \;power}{Noise\;power}
\end{equation}

The definitions of signal power and noise power are as follows:  
\begin{definition}
Signal power refers to the power of the desired signal, which is the information or data being transmitted or received. Mathematically, it is calculated as the average or mean squared value of the signal. \newline
In the case of a discrete signal ($s[n]$), which has values for only discrete points in time, the signal power $P_{s}$ is represented as follows:
$N$: The number of samples taken for computation from a snapshot of the signal over an arbitrary time duration,
\begin{equation}
P_{s} = lim_{N \to \infty} \frac{1}{2N+1} \sum_{n=-N}^{N} |s[n]|^2
\end{equation}
\end{definition}

\begin{definition}
Noise power represents the power of the unwanted signal or interference, which corrupts the desired signal. Similar to signal power, noise power is often calculated as the average or mean squared value of the noise. \newline
Similar to the signal power, for a discrete noise $n[n]$, the noise power is represented as:  
\begin{equation}
P_{n} = lim_{N \to \infty} \frac{1}{2N+1} \sum_{n=-N}^{N} |n[n]|^2
\end{equation}
\end{definition}

While signal power measures the "strength" or "magnitude" of the signal. Signal variance (denoted as $\sigma^2_{s}$) measures how much the signal values deviate from the mean ($\mu_{s}$). It provides an indication of the "spread" or "dispersion" of the signal values around their average.
In the general case, the relationship between the variance and power for a signal with a non-zero mean is: 
\begin{equation}
    \sigma^2_{s} = P_{s} - \mu_{s}^2
\end{equation}
 The signal variance for discrete Signals:
\begin{equation}
    \sigma^2_{s} = \frac{1}{N} \sum_{n=1}^{N} (s[n] - \mu_{s})^2
\end{equation}
In the case in which the mean of the signal is zero, the power is equivalent to the signal variance. The same computations can be applied to the noisy signal. 

SNR, often expressed in decibels, is sensitive to the scale of the noise and signal in the system. Higher SNR indicates the signal is of high quality and is easier to identify from noise. Conversely, when SNR is low, the signal is weak, or the system is too noisy, and distinguishing the true signal from noise is more challenging. We redefine SNR using the signal variance and noise variance as: 
\begin{equation}
SNR = 10\times \log_{10}\frac{Signal \;variance}{Noise\;variance}    
\end{equation}

SNR is used as a metric to evaluate the strength of the signal in the presence of noise and achieve optimal performance. In this context, using signal variance over signal power offers certain advantages: 
\begin{itemize}
\item Variance captures the fluctuations of the signal around its mean. In the case of CNN models, the variance provides an understanding of the model's confidence or consistency in its responses. By focusing on variance, the model's behavior is tied directly to the properties of noise. A higher noise variance indicates that the model is more uncertain and less stable in the presence of noise.
\item Variance is a normalized measure, making it a relative metric. This can be advantageous when comparing the performance of different models or the same model under varying noise conditions, as it ensures that the measure is scaled and comparable.
\end{itemize}

Computing SNR based on the model output is a quantitative tool for evaluating the model's performance in detecting useful information (signal) from unwanted variations (noise) in the data. SNR allows us to observe the changes in the output and find the noise level that meets the desired trade-off between accuracy, stability, and generalization. Understanding the impact of noise during training provides a guideline for determining the privacy budget without concerns about the quality of results. Leveraging noise to improve stability and generalization without sacrificing performance leads to stronger privacy protection strategies against adversaries.

In the context of CNN, the signal represents the true underlying patterns that the model is trying to capture, and noise is any internal or external variation, perturbation, or distortion in the data that affects the model's ability to detect the signal. The formal definition of signal and noise is provided:
\begin{definition}
A signal is the validation accuracy of the base model (model without noise). 
\end{definition}

\begin{definition}
Noise is defined as the difference between the base model's validation accuracy and the perturbed model's validation accuracy. A model is perturbed by introducing a randomly generated value from the Gaussian distribution with a standard deviation of $\sigma$.
\end{definition}

Using validation accuracy obtained from the noisy and clean data provides a more reliable assessment of how well a model handles noise and generalizes to new, challenging conditions. Training accuracy tends to overstate performance, while test accuracy is reserved for final evaluation and should not be influenced by noise during model development.

The choice of the noise infusion scenario relies on the problem's complexity and dataset. In classification, higher SNR values indicate that the model is capable of predicting values that are closer to the true signal and have less noise interference. The lower SNR values suggest that the noise is more dominant, resulting in less accurate predictions by the model. The noise level that yields the maximum SNR is preferable because it identifies the noise level where the model can most extract useful information from noise, leading to better generalization of unseen data. 
\subsection{Price of Stability \& Price of Anarchy}
Originally used for the analysis of network and routing games, the Price of Stability (PoS) and Price of Anarchy (PoA) measure the efficiency of outcomes in decentralized systems \cite{anshelevich2008price, koutsoupias2009worst}. PoS compares the outcome achieved by self-interested agents to the socially optimal solution. PoA compares the worst-case outcome achieved by self-interested agents to the socially optimal solution. We propose to define the image classification process as a game where the players are Gaussian noise-infused CNNs under various noise levels. For $N$ players, and $i = 1, ..., n$, the standard deviation of the Gaussian noise of the $i^{th}$ player is $\sigma_{i}$,  where $\sigma_{i} \in [0, 1]$. Suppose the ideal scenario is training the model without noise (base model denoted as $CNN_{\sigma_{0}}$). PoS is defined as the ratio of the model with noise and the base model:
\begin{equation}
\begin{split}
Price\; of\; Stability\;(PoS_{i}) = \frac{Test\;accuracy\;of\;CNN_{\sigma_{i}}}{Test\;accuracy\;of\;CNN_{\sigma_{0}}}
\end{split}
\end{equation}

By comparing against the base model, we can assess how training with noise impacts the prediction results of the test data. 
\begin{itemize}
    \item The PoS of the base model is always 1. 
    \item If PoS $=$ 1, the model's sensitivity to noise is minimal. The noisy model is performing similarly to the base model. It also suggests that the model is relatively stable across different noise levels.
    \item  If PoS $>$ 1, the noisy model performs better than the base model. It suggests that additive noise improves the model's generalization on unseen data. Therefore, test accuracy has improved in the presence of noise.
    \item If PoS $<$ 1, the noisy model performs worse than the base model. Smaller PoS suggests a lack of stability in the presence of noise. The model has less potential for privacy-preserving applications.
\end{itemize}
The PoA is defined as:
\begin{equation}
\begin{split}
Price\; of\; Anarchy\;(PoA_{i}) = \frac{Test\;loss\;of\;CNN_{\sigma_{i}}}{Test\;loss\;of\;CNN_{\sigma_{0}}}
\end{split}
\end{equation}

\begin{itemize}
\item The PoA of the base model is always 1.
\item If PoA $=$ 1, the model is able to identify the patterns in the data, even in noisy conditions.
\item   If PoA $>$ 1, the model is negatively impacted by the noise, and it loses useful information, so the noisy model performs worse than the base model.
\item  If PoA $<$ 1, the model performs better than the base model, and the additive noise has improved the model's generalization on unseen data.
\end{itemize}

The proposed metrics provide insights into the effect of noise on the models' accuracy, loss, and overall stability. The metrics also offer a clear reference point to monitor the changes in the models' generalization and efficiency of predictions on test data.
\section{Computational Results and Analysis}\label{sec-result}
In this section, we explore the use of noise as a means of improving generalization, stability, and privacy in deep neural networks. This is particularly important when data is distributed across multiple devices and access to sufficient data for training is limited. We aim to design stable and differentially private deep learning models that can generalize well in centralized and federated learning settings while preserving privacy. To achieve this goal, we will compare various methods of designing algorithms that can perform well in the presence of noise and evaluate their effectiveness for image classification. We will build upon the foundational work of Zhang et al. \cite{zhang2021understanding} and expand their findings through our experimentation.

We start the experiments by selecting the appropriate CNN architecture. As mentioned earlier, the VC dimension is the measure of the model's expressive power and is often used to analyze the model's capacity to fit data. Training large CNN models with millions of trainable parameters requires significant computation resources and careful fine-tuning of the hyperparameters. We use CIFAR-10, a well-known benchmark dataset for image classification, where 40,000 images are used for training, 10,000 images for validation, and 10,000 for testing. Our experiments are designed around three network architectures with different model capacities determined by the number of parameters in the neural network architecture provided in Table \ref{table-Architecture}.
\begin{table*}[ht]
\centering
\caption{\label{table-Architecture}The models vary in the number of trainable parameters, a factor of model capacity that impacts the model's ability to generalize on unseen data. Model 3 is over-parameterized}
\begin{tabular}{|l|c|c|c|}
\hline
Architecture & Trainable Param & Non-trainable Param & Total \\
\hline
Model 1 & 22,784,938 & 1,920  & 22,786,858 \\
Model 2 & 2,396,330 & 1,896 & 2,397,226 \\
Model 3 & 43,415,850 & 3,968 & 43,411,882 \\
\hline
\end{tabular}
\end{table*}

The CNN models are modifications of VGG-19 \cite{simonyan2014very}, and the key layers are the 2D convolutional, batch normalization, 2D max pooling, dropout, and dense layers. The architecture details for models 1, 2, and 3 are available in Tables \ref{table-Model1}, \ref{table-Model2}, and \ref{table-Model3}, respectively. The parameters of the CNN are configured as a batch size of 64, a learning rate of 0.001, and a momentum of 0.9. The local models are trained for 80 epochs. The three models with different numbers of parameters are compared in their efficiency of prediction, generalization, and stability under different noise levels and noise infusion mechanisms.  
\begin{table}[ht]
\centering
\caption{\label{table-Model1}Architecture for Model 1}
\begin{tabular}{|l|c|c|}
\hline
Layer Type              & Output Shape  \\
\hline
Conv2D                  & (32, 32, 32) \\
BatchNormalization      & (32, 32, 32) \\
Conv2D                  & (32, 32, 32) \\
BatchNormalization      & (32, 32, 32) \\
MaxPooling2D            & (16, 16, 32) \\
Dropout                 & (16, 16, 32) \\
Conv2D                  & (16, 16, 64) \\
BatchNormalization      & (16, 16, 64) \\
Conv2D                  & (16, 16, 64) \\
BatchNormalization      & (16, 16, 64) \\
MaxPooling2D            & (8, 8, 64)   \\
Dropout                 & (8, 8, 64)   \\
Conv2D                  & (8, 8, 128)  \\
BatchNormalization      & (8, 8, 128)  \\
Conv2D                  & (8, 8, 128)  \\
BatchNormalization      & (8, 8, 128)  \\
MaxPooling2D            & (4, 4, 128)  \\
Dropout                 & (4, 4, 128)  \\
Conv2D                  & (4, 4, 256)  \\
BatchNormalization      & (4, 4, 256)  \\
Conv2D                  & (4, 4, 256)  \\
BatchNormalization      & (4, 4, 256)  \\
Conv2D                  & (4, 4, 256)  \\
MaxPooling2D            & (2, 2, 256)  \\
Dropout                 & (2, 2, 256)  \\
Flatten                 & (1024,)      \\
Dense                   & (4096,)      \\
Dropout                 & (4096,)      \\
Dense                   & (4096,)      \\
Dense                   & (10,)        \\
\hline
\end{tabular}
\end{table}

\begin{table}[ht]
\centering
\caption{\label{table-Model2}Architecture for Model 2}
\begin{tabular}{|l|c|c|}
\hline
Layer Type              & Output Shape \\
\hline
Conv2D                  & (32, 32, 32) \\
BatchNormalization      & (32, 32, 32) \\
Conv2D                  & (32, 32, 32) \\
BatchNormalization      & (32, 32, 32) \\
MaxPooling2D            & (16, 16, 32) \\
Conv2D                  & (16, 16, 64) \\
BatchNormalization      & (16, 16, 64) \\
Conv2D                  & (16, 16, 64) \\
BatchNormalization      & (16, 16, 64) \\
MaxPooling2D            & (8, 8, 64)   \\
Conv2D                  & (8, 8, 128)  \\
BatchNormalization      & (8, 8, 128)  \\
Conv2D                  & (8, 8, 128)  \\
BatchNormalization      & (8, 8, 128)  \\
MaxPooling2D            & (4, 4, 128)  \\
Flatten                 & (2048,)      \\
Dropout                 & (2048,)      \\
Dense                   & (1024,)      \\
Dropout                 & (1024,)      \\
Dense                   & (10,)        \\
\hline
\end{tabular}
\end{table}

\begin{table}[t]
\centering
\caption{\label{table-Model3}Architecture for Model 3}
\begin{tabular}{|l|c|}
\hline
Layer Type              & Output shape  \\
\hline
Conv2D                  & (32, 32, 32) \\
BatchNormalization      & (32, 32, 32) \\
Conv2D                  & (32, 32, 32) \\
BatchNormalization      & (32, 32, 32) \\
MaxPooling2D            & (16, 16, 32) \\
Conv2D                  & (16, 16, 64) \\
BatchNormalization      & (16, 16, 64) \\
Conv2D                  & (16, 16, 64) \\
BatchNormalization      & (16, 16, 64) \\
MaxPooling2D            & (8, 8, 64)   \\
Conv2D                  & (8, 8, 128)  \\
BatchNormalization      & (8, 8, 128)  \\
Conv2D                  & (8, 8, 128)  \\
BatchNormalization      & (8, 8, 128)  \\
MaxPooling2D            & (4, 4, 128)  \\
Conv2D                  & (4, 4, 256)  \\
BatchNormalization      & (4, 4, 256)  \\
Conv2D                  & (4, 4, 256)  \\
BatchNormalization      & (4, 4, 256)  \\
Conv2D                  & (4, 4, 256)  \\
MaxPooling2D            & (2, 2, 256)  \\
Dropout                 & (2, 2, 256)  \\
Conv2D                  & (2, 2, 512)  \\
BatchNormalization      & (2, 2, 512)  \\
Conv2D                  & (2, 2, 512)  \\
BatchNormalization      & (2, 2, 512)  \\
Conv2D                  & (2, 2, 512)  \\
MaxPooling2D            & (1, 1, 512)  \\
Dropout                 & (1, 1, 512)  \\
Flatten                 & (512,)       \\
Dense                   & (4096,)      \\
Dropout                 & (4096,)      \\
Dense                   & (8192,)      \\
Dense                   & (10,)        \\
\hline
\end{tabular}
\end{table}

\subsection*{Experiment 1: CNN with Gaussian noise hidden layers in Centralized Setting}
\begin{figure*}[ht]
\centering
\includegraphics[width = 1 \linewidth, keepaspectratio]{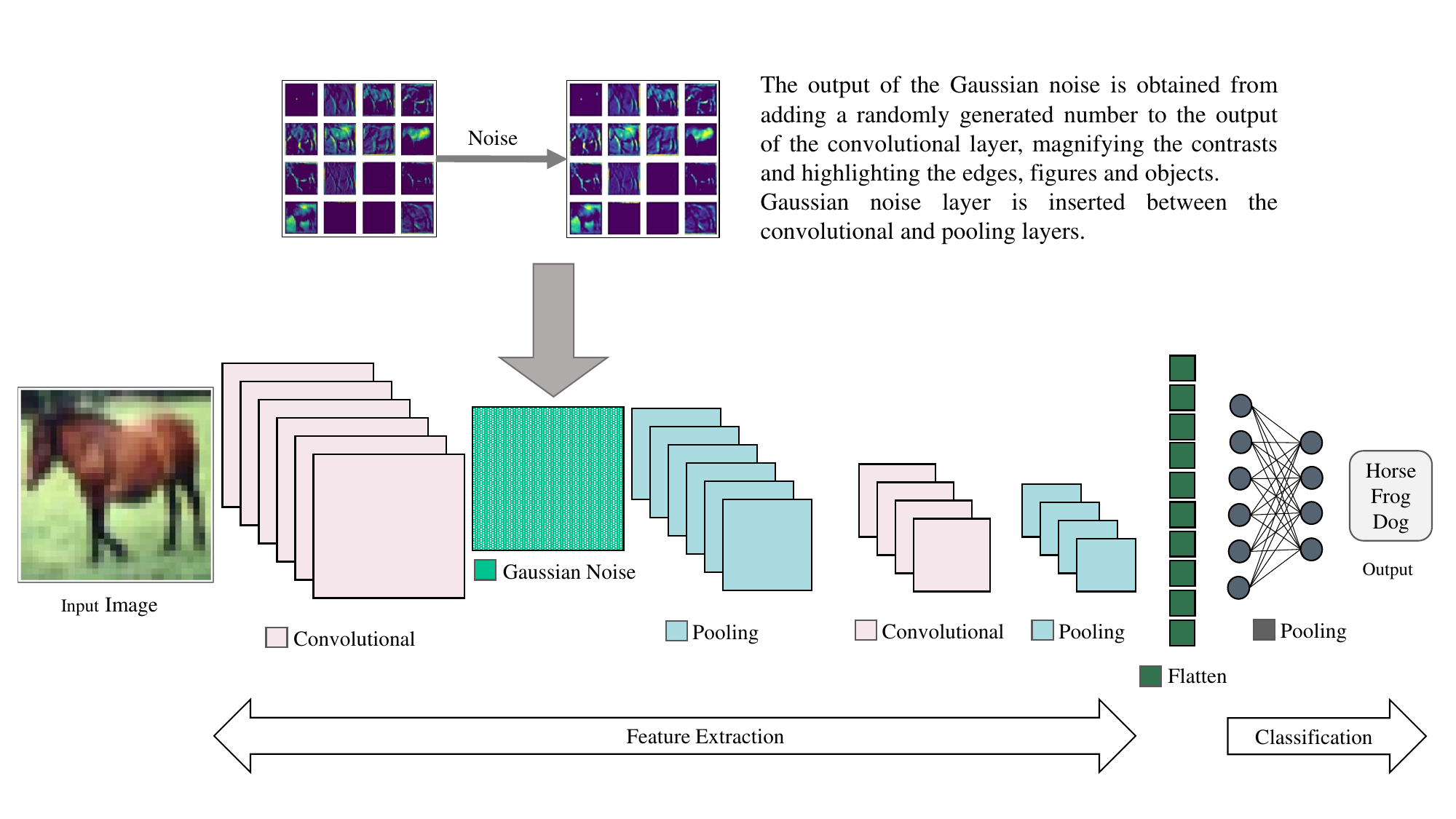}
\caption{A simplified illustration of the CNN architecture with Gaussian noise layer.}
\label{CNN}
\end{figure*}

Leveraging the properties of training with noise, we design a CNN with Gaussian noise hidden layers, an innovative approach to enhance the robustness and generalization capabilities of deep learning models. In this design illustrated in Figure\ref{CNN}, Gaussian noise is intentionally added as a form of regularization to hidden layers within the CNN architecture. 

Training with Gaussian noise hidden layers involves inserting uncorrelated layers of Gaussian noise that will add a randomly generated value within the range of the specified standard deviation to the activation of the previous layer during training. Uncorrelated noise sources are statistically independent.

\begin{figure*}[ht]
    \centering
    \includegraphics[width = 1 \linewidth, keepaspectratio]{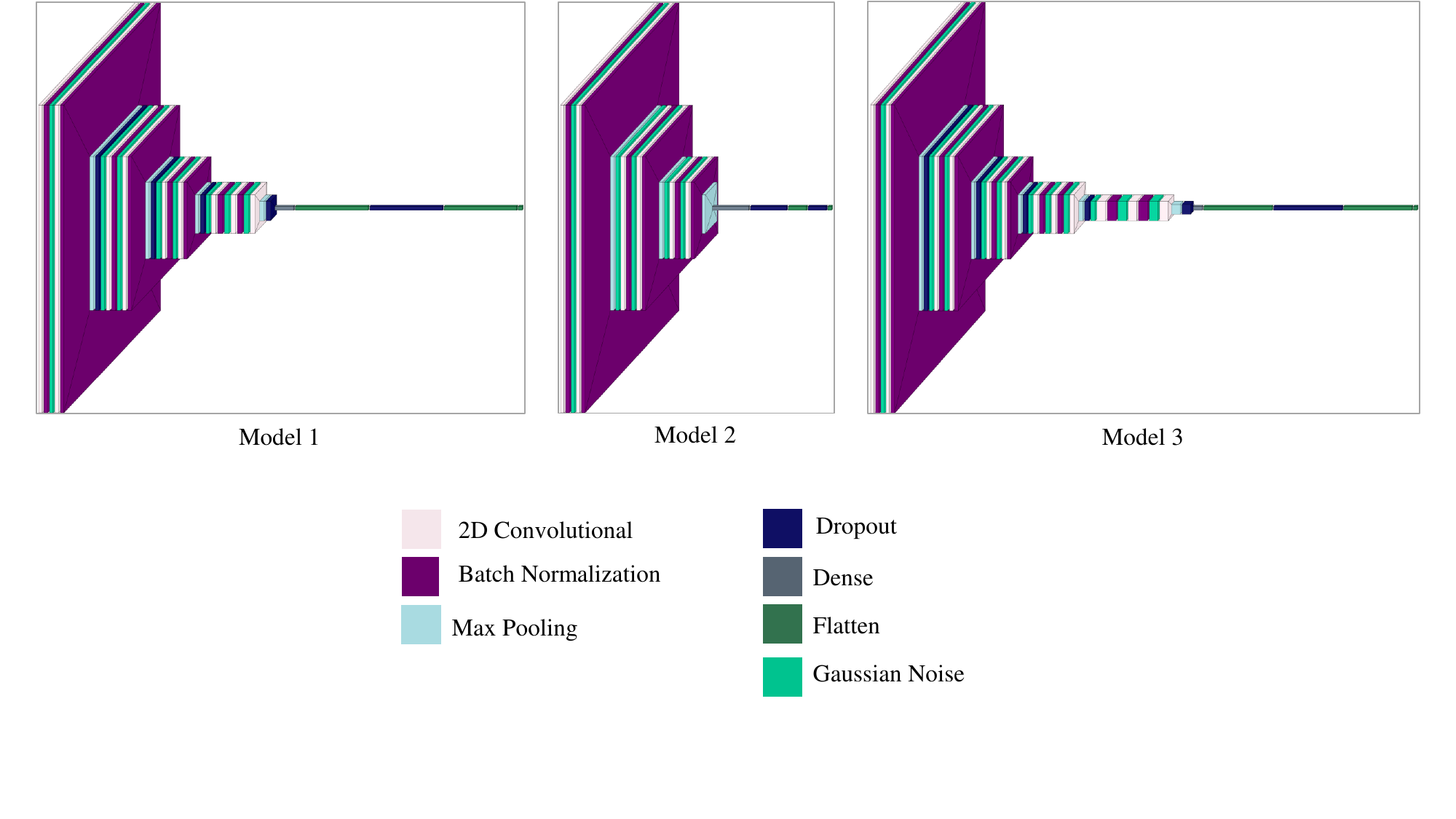}
    \caption{Visual representation of the CNN models with Gaussian noise layers.}
    \label{Noisy architectures}
\end{figure*}

In the first set of experiments, we evaluate the performance of three CNN models with Gaussian noise hidden layers presented in Figure \ref{Noisy architectures}.  

For the implementation of the models in this study, we insert the noise layers before the convolutional layers, followed by a batch normalization layer.
Let us assume: \newline
$x$: The output of the layer before the convolutional layer \newline
$z$: A randomly generated number from Gaussian distribution with mean, $\mu = 0$ and standard deviation of $\sigma$ \newline
$x\prime$: The output of the Gaussian noise layer, $x\prime = x + z$. \newline
$x\prime\prime$: The output of the batch normalization layer obtained after passing the output of the convolutional layer through a batch normalization layer:
\begin{equation}
    x\prime\prime = (\frac{x\prime - \mu\prime}{\sigma\prime})*\alpha + \beta
\end{equation}
Where $\mu\prime$ and $\sigma\prime$ are the mean and standard deviation of the neuron's output of the activation function in the convolutional layer, and $\alpha$ and $\beta$ are trainable parameters used for rescaling and shifting the values from the previous operations. As the training continues, the data goes through multiple blocks of Gaussian noise, convolutional, and batch normalization layers. Batch normalization prevents the accumulation of noise throughout the network.

\begin{figure*}[ht]
\centering
\includegraphics[width = 1 \linewidth, keepaspectratio]{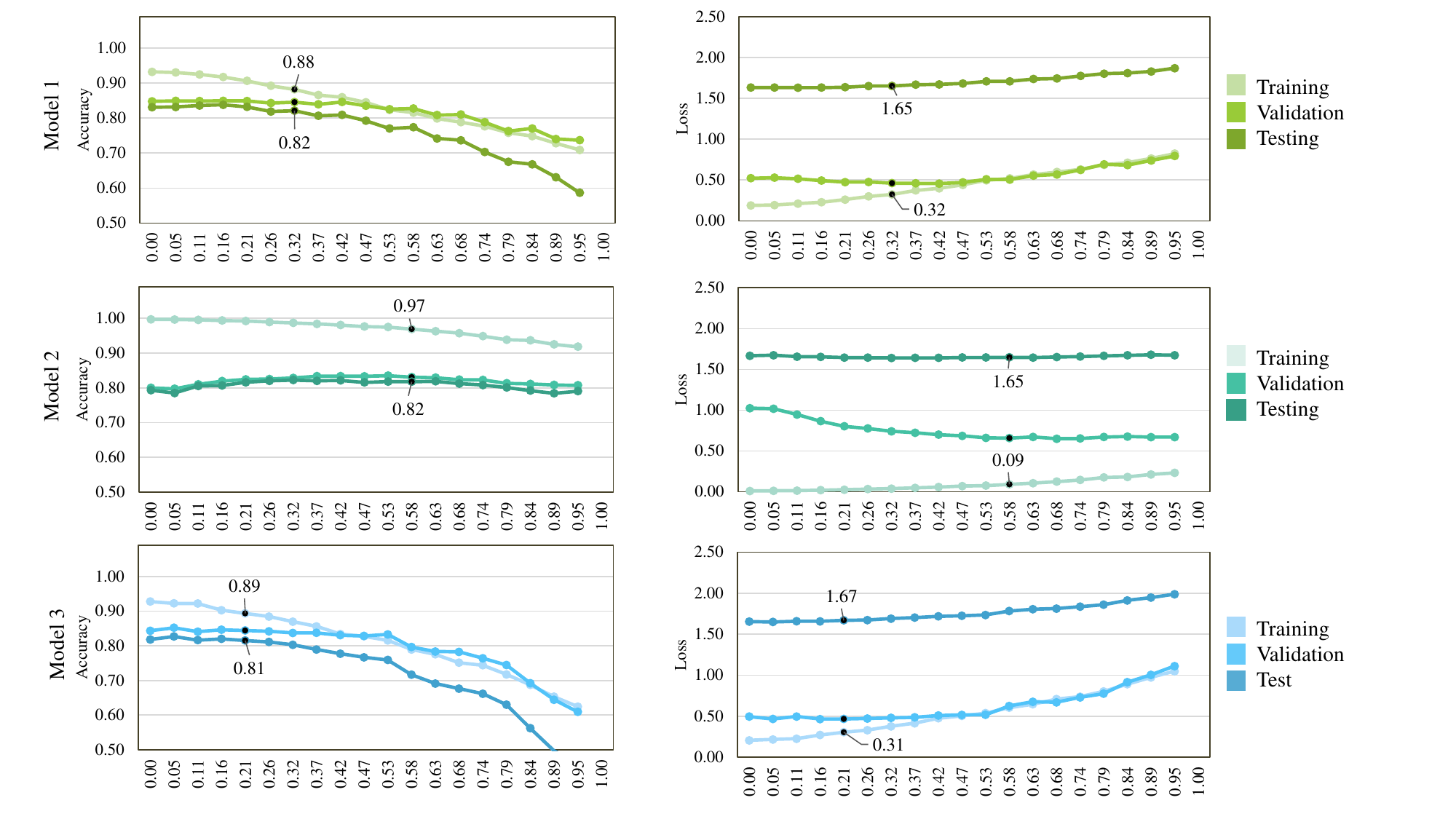}
\caption{The optimal test accuracy and loss value are marked with the associated training accuracy and loss. Stable models that perform well at higher noise levels are better candidates for federated learning. }
\label{Model comparison-1}
\end{figure*}

Figure \ref{Model comparison-1} compares accuracy and loss obtained from training the models under different noise levels in a centralized framework. The standard deviation is selected from Gaussian distribution with 20 levels between $\{0, 1\}$. Setting the standard deviation to zero refers to the base model. 

Models 1 and 3 offer similar trends; as noise increases, the accuracy drops, and loss increases further from the base model. In models 1 and 3, the optimal test accuracy and loss are achieved when $\sigma$ are 0.32 and 0.21, respectively. The drop in performance as a result of increasing the noise suggests that the models have difficulty fitting the noisy data when $\sigma$ is high. 

Unlike models 1 and 3, model 2 can maintain consistent performance with noisy data, suggesting that the model is the most stable among the three. In model 2, the optimal test accuracy and loss are achieved when $\sigma$ is 0.58, which is significantly higher than in models 1 and 3. While all three models yield the optimal accuracy of approximately 0.82, maintaining a high accuracy and loss in the presence of higher noise levels demonstrates that model 2 is better at generalizing to unseen data. Compared to models 1 and 3, model 2 experiences a less rapid performance degradation at higher noise levels. 

\begin{figure*}[ht]
    \centering
    \includegraphics[width = 1 \linewidth, keepaspectratio]{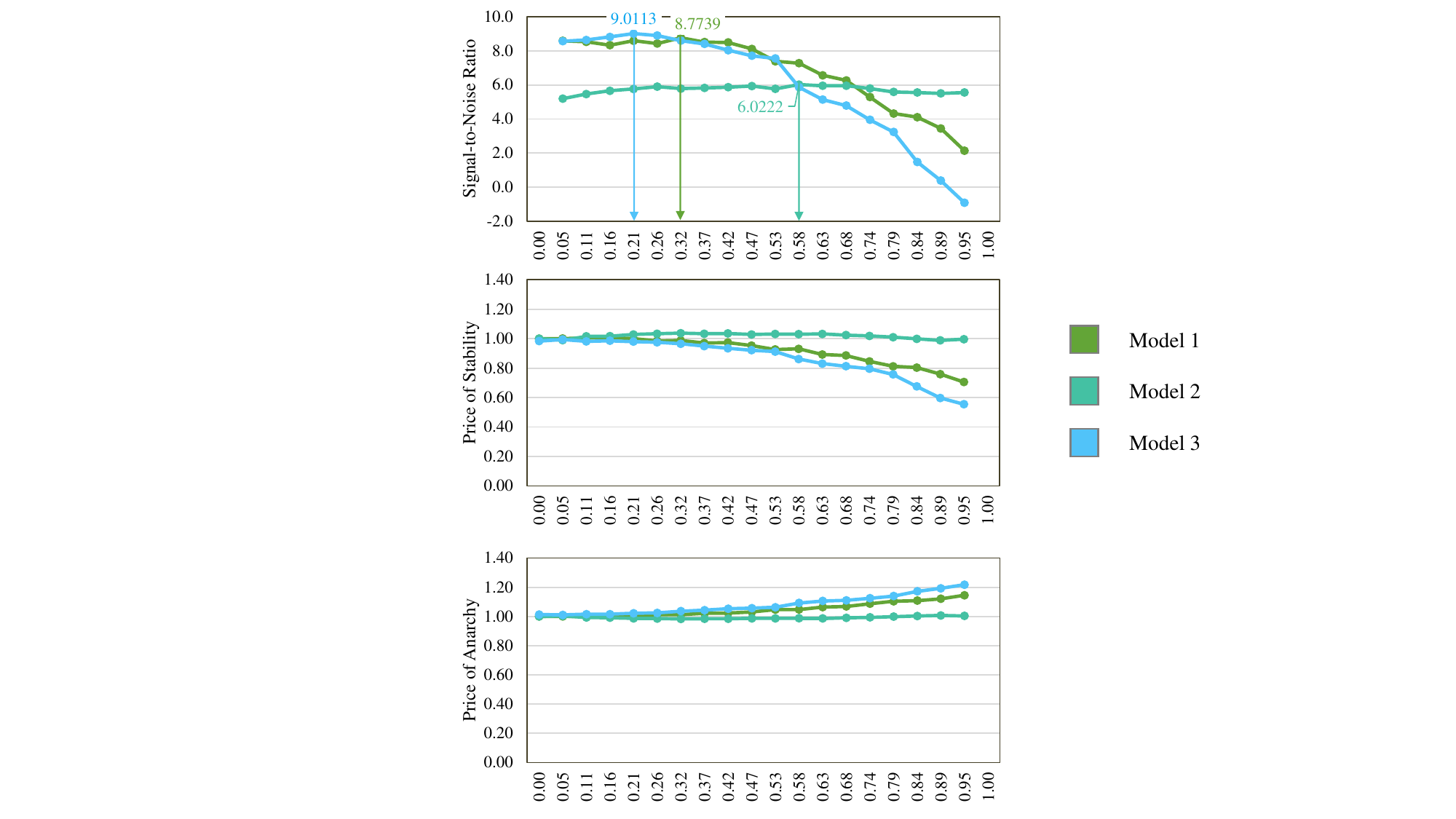}
    \caption{Increasing the noise levels decreases model utility. However, stable models suffer less as the noise levels are heightened, offering consistent performance under higher noise levels.}
    \label{Model comparison-2}
\end{figure*}

Often, better privacy guarantees are achieved at the expense of worse accuracy and loss, so we strive to find a systematic way to reach a balance between accuracy and privacy. However, the balance is not possible without fine-tuning the noise level during training while monitoring its impact on test data. To this end, we explore SNR, PoS, and PoA to measure the trade-off between performance efficiency and privacy under noise. 
Figure \ref{Model comparison-2} demonstrates the SNR, PoS, and PoA values for the three models with Gaussian noise hidden layers ($\sigma$  between 0 and 1).

Since the range of SNR is problem-dependent, we focus on the fluctuations of SNR at different noise levels to compare the models. 

In models 1 and 3, the value of SNR is initially higher but drops significantly as we increase the noise. This means at lower noise levels, the model is effective in distinguishing the signal, but as noise increases, the model becomes overwhelmed and can not handle noise effectively. However, model 2 stands out as having relatively consistent SNR values at higher noise levels.  This implies the model's ability to remain relatively stable, even in the presence of higher noise levels. 

Training the models at the noise level that maximizes SNR provides the highest test accuracy and sets the balance between stability and accuracy in the presence of noise. Under differential privacy, the maximum SNR guarantees privacy without loss of accuracy. While finding the balance is ideal, in federated learning, privacy is prioritized over accuracy when dealing with sensitive data. PoS and PoA measure the impact of noise on test accuracy and loss compared to the base model. In Model 2, the PoS and PoA remain consistent despite the increase in the noise level. Model 2 offers a trade-off between performance and privacy, where accuracy and loss are stable under higher noise levels. In model 2, while the optimal SNR identifies the noise level for the perfect balance between accuracy and privacy at 0.58, we can further increase the noise, and the accuracy degrades by less than 4$\%$. Model 2 is a potential candidate for cases where privacy and stability take precedence over achieving the highest accuracy, such as federated learning applications. 

Ultimately, selecting the appropriate model depends on the specifics and requirements of the problem, whether it prioritizes accuracy, privacy, or stability. These analyses provide insights into the trade-offs and strengths of each model under different noise levels. 

Overall, a comparison of the performance of the three models under various noise conditions measured by SNR, PoS, and PoA suggests that in models with higher stability, PoS and PoA remain relatively consistent. Given the overlap between the definitions of stability and privacy, we can conclude that models with relatively consistent PoS and PoA can provide better privacy protection guarantees without drastic degradation of accuracy. 

\subsection*{Experiment 2: CNN with Multiple Gaussian Noise Layers vs. a Single Layer}
\begin{figure*}[ht]
\centering
\includegraphics[width = 1 \linewidth, keepaspectratio]{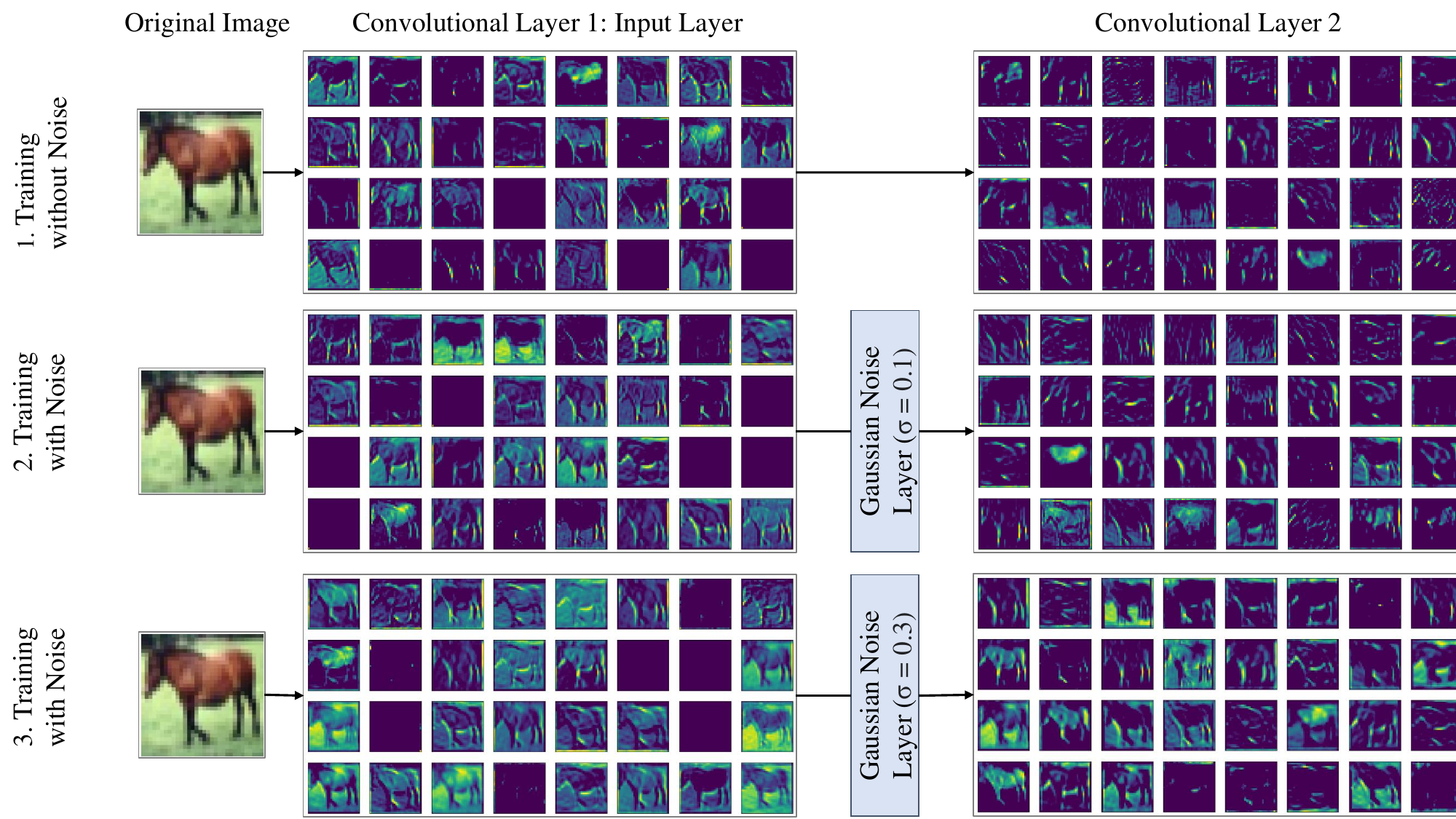}
\caption{CNN feature maps}
\label{Feature map-1}
\end{figure*}

When an image is passed through the convolutional layers, the network learns different complex features of the image, such as the edges and the texture. The network learns patterns and objects from the later convolutional layers as training continues. We use feature visualization to gain insight into the learning procedure of a CNN with Gaussian noise hidden layers inserted before the convolutional layer, focusing on the first layers of model 1. Figure \ref{Feature map-1} is a visual representation of the output of the first two convolutional layers of model 1, where a single image is fed into the network. 

The first two convolutional layers have 32 filters. The figure includes three sets of feature maps from the initial training steps extracted from the model without noise and the noisy model, where a noise layer is inserted before the second convolutional layer. The first column represents the feature maps from the input layer of CNN models. The slight variations in the maps are due to the inherent variations in training a neural network model. The lower layers of the CNN are responsible for learning the edges and textures in the image. The bright spots on the feature map indicate that the region was most activated in its corresponding map in the prior layer. 

\begin{figure*}[ht]
\centering
\includegraphics[width = 1 \linewidth, keepaspectratio]{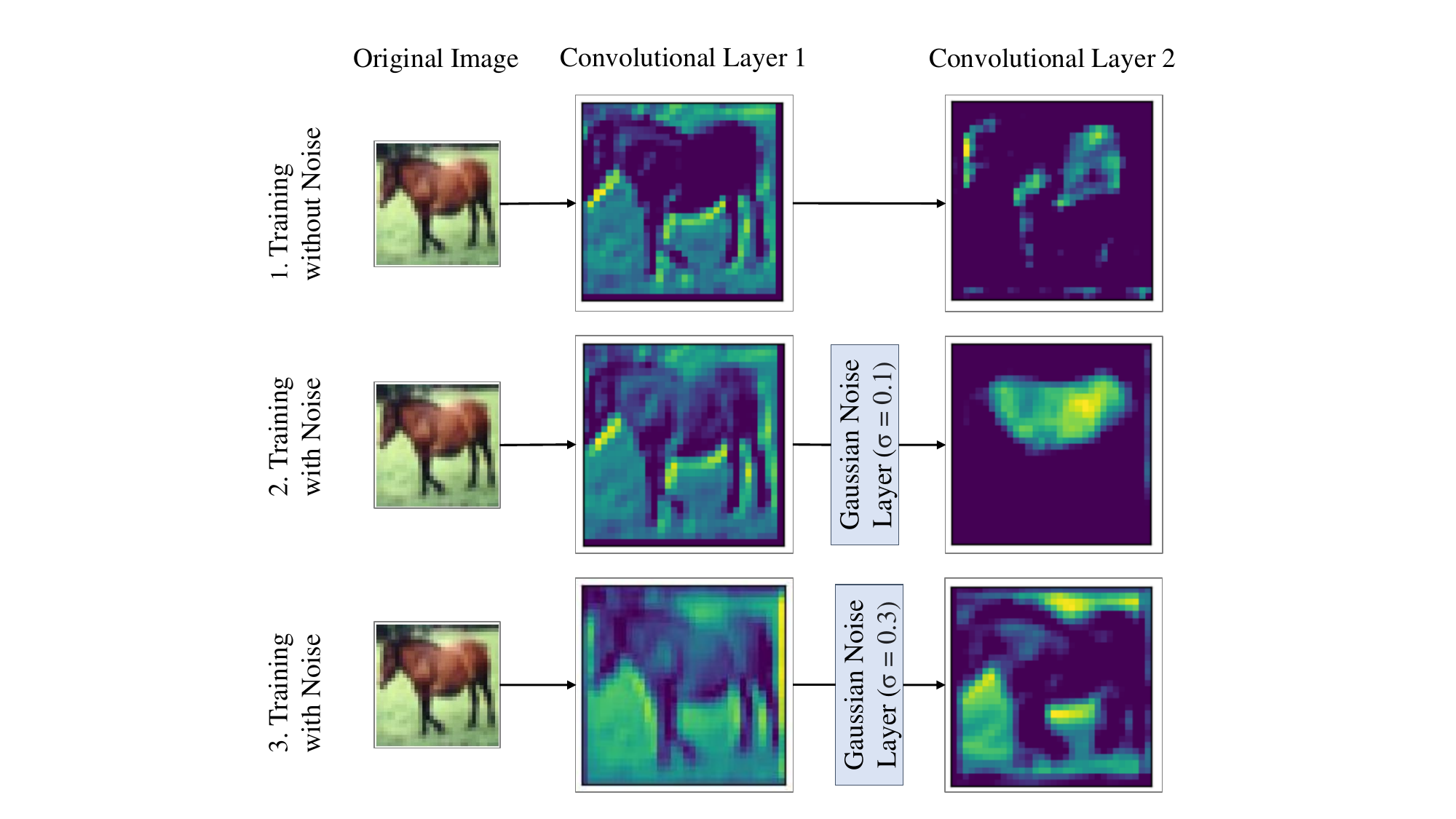}
\caption{The optimal noise level improves generalization by helping the deep learning model better distinguish the objects during training.}
\label{Feature map-2}
\end{figure*}

In training with noise, we utilize the idea of stochastic resonance and use noise to enhance weak signals. Figure \ref{Feature map-2} is a closer look at the feature map. For a relatively similar map in layer 1, the noise-infused maps in the second and third rows have led to better identification of edges, and more key regions are activated. 

We emphasize that uncorrelated noise sources are critical when designing CNN models with Gaussian noise hidden layers. If noise layers are correlated, we must consider different magnitudes and phase variations when combining the additive noise. We can ensure that the noise layers are uncorrelated by assigning a unique random seed at each layer. Derived from the signal processing conventions, we can compute the total additive noise in the system for multiple statistically independent noise sources. 

Let us assume that $G_{1}$ and $G_{2}$ are two uncorrelated noise layers, with standard deviations $\sigma_{1}$ and $\sigma_{2}$, respectively.
\begin{equation}
\begin{split}
Variance(G_{1}, G_{2}) = Variance(G_{1}) + Variance(G_{2}) + 2\rho*CoVariance(G_{1}, G_{2})
\end{split}
\end{equation}\label{uncorrelated}
Since the noise layers are uncorrelated, $\rho = 0$, and, 
\begin{equation}
    Variance(G_{1} + G_{2}) = \sigma_{1}^2 + \sigma_{2}^2.
\end{equation}\label{stddev}

In a CNN model with N uncorrelated Gaussian noise hidden layers, the total noise introduced by multiple noise layers with standard deviation $\sigma$ is equivalent to a single noise layer with a standard deviation of: 
\begin{equation}
  \sigma_{Total} = \sqrt{N}\sigma  
\end{equation}

Table \ref{sigle noise layer} presents the results from training models with multiple noisy layers vs. a single noise layer. For comparison, we set $\sigma = 0.1$ when training all three models with multiple noise layers shown in Figure \ref{Noisy architectures}.
The standard deviation of the model with a single noise layer is computed based on Equation \ref{stddev}. 
\begin{table*}[ht]
\caption{Training deep learning models with a single Gaussian noise hidden layer versus multiple layers.}
\centering
\begin{tabular}{|l|l|l|l|}
\hline
\multicolumn{4}{|c|}{Model 1}\\
\hline
 Model Type & Standard Deviation & Train Accuracy & Test Accuracy\\ 
\hline
Base model & 0.0 & 0.93 & 0.82\\
Multiple noise layers & 0.1 & 0.92 & 0.83\\
Single layer substitute & 0.28 & 0.92 & 0.83\\
\hline
\hline
\multicolumn{4}{|c|}{Model 2}\\
\hline
 Model Type & Standard Deviation & Train Accuracy & Test Accuracy\\ 
\hline
Base model & 0.0 & 0.99 & 0.79\\
Multiple noise layers & 0.1 & 0.99 & 0.80\\
Single layer substitute & 0.22 & 0.99 & 0.80\\
\hline
\hline
\multicolumn{4}{|c|}{Model 3}\\
\hline
 Model Type & Standard Deviation & Train Accuracy & Test Accuracy\\ 
\hline
Base model & 0.0 & 0.92 & 0.82\\
Multiple noise layers & 0.1 & 0.91 & 0.83\\
Single layer substitute & 0.33 & 0.92 & 0.83\\
\hline
\end{tabular}
\label{sigle noise layer}
\end{table*}

The obtained results suggest that the number of layers does not affect the model performance. In this framework, the controlling parameter is the standard deviation of the added noise. Training the models with multiple noise layers allows us to fine-tune the standard deviation of the noise generated at the layers and adjust the model according to the problem specifications and data at hand to achieve optimal performance. 

\subsection*{Experiment 3: Gaussian noise hidden layers in Federated Setting }
Extending the experiments to federated learning, we explore the effect of different noise levels and compare the results with the centralized models. Choosing the Gaussian noise magnitude is critical because it determines the level of privacy. A lower noise level will result in a more accurate CNN model but will also provide weaker privacy guarantees. 
It is important to note that, in practice, it is possible for an attacker to learn sensitive information about the training data by exploiting vulnerabilities in the model or the training process. Therefore, it is important to take additional steps to protect the privacy of the training data, such as using secure training environments and encryption. Using horizontal partitioning, the data is randomly and equally split between 3 arbitrary clients. 

First, the models were trained locally with 20 noise levels between $\{0, 1\}$, and SNR was computed. The noise level that yields the optimal SNR for the clients and the results from training the federated learning model with optimized noise obtained from maximizing SNR are presented in Table \ref{Clients-SNR-Federated}. The federated learning models are trained for 20 communication rounds at different noise levels. Global accuracy and global loss are measured for evaluation. It can be observed that despite significant differences in size, the models vary by a maximum of 3\% in global accuracy, while global loss remains relatively consistent.
\begin{table*}[ht]
\caption{The standard deviation of the additive noise is set based on the optimal SNR.}
\centering
\begin{tabular}{|l|l|l|l|l|l|}
\hline
Architecture & Client 1 & Client 2  & Client 3  & Global Loss& Global Accuracy\\ 
\hline
Model 1 & 0.21 & 0.26 & 0.16 & 1.60& 0.87\\ 
Model 2 & 0.53 & 0.37 & 0.16 & 1.63& 0.84\\
Model 3 & 0.11 & 0.16 & 0.16 & 1.60& 0.86\\
\hline
\end{tabular}
\label{Clients-SNR-Federated}
\end{table*}

In the next step, we trained the model at five noise levels, the results of which are shown in \ref{noise layers-Federated}. Training the models with Gaussian noise hidden layers significantly improves the model stability.  

As seen in Figure \ref{Model comparison-2}, it is possible to add higher noise levels to improve privacy guarantees, and the global accuracy and loss remain relatively constant with varying noise levels.
\begin{table*}[ht]
\caption{The standard deviation of the additive noise is fixed across all clients.}
\centering
\begin{tabular}{|l|l|l|l|l|l|}
\hline
\multicolumn{6}{|c|}{Global Accuracy}\\
\hline
Architecture & $\sigma = $0.1& $\sigma = $0.3& $\sigma = $0.5& $\sigma = $0.7& $\sigma = $0.9\\ 
\hline
Model 1 & 0.86& 0.86& 0.87& 0.86& 0.86\\
Model 2 & 0.78& 0.83& 0.84& 0.84& 0.86\\ 
Model 3 & 0.85& 0.86& 0.87& 0.87& 0.85\\
\hline
\hline
\multicolumn{6}{|c|}{Global Loss}\\
\hline
Architecture & $\sigma = $0.1& $\sigma = $0.3& $\sigma = $0.5& $\sigma = $0.7& $\sigma = $0.9\\ 
\hline
Model 1 & 1.60& 1.61& 1.60& 1.61& 1.62\\
Model 2 & 1.68& 1.64& 1.63& 1.63& 1.62\\ 
Model 3 & 1.61& 1.61& 1.61& 1.63& 1.72\\
\hline
\end{tabular}
\label{noise layers-Federated}
\end{table*}

The analysis suggests that deep learning models are relatively noise-stable in federated settings. The models can learn the patterns of the data and the added noise while preserving privacy. The stability of the models in federated learning is beneficial as it increases the model's threshold for added noise, ensuring that privacy is maintained. Increasing the standard deviation of Gaussian noise, which acts as a regularization method, also improves the overall accuracy of test data. 
\subsection*{Experiment 4: Comparison of Noise Infusion Mechanisms}
The choice of noise infusion mechanism plays a crucial role in enhancing deep learning models' generalization, stability, and privacy. This section compares the impact of noise infusion schemes mentioned in Table \ref{literature}.
\begin{itemize}
    \item Noisy input: The input noise is implemented by adding a random value sampled from the Gaussian distribution in the predefined standard deviation range to the input data during training. Input noise behaves as a data augmentation method, often used to expand the input sample or introduce randomness in the data to reduce overfitting. However, if the noise level is too high, it can distort the data and lead to the model learning incorrect patterns.
    \item Noisy network weights: To introduce noise to model weights, the noise is directly added to the weights retrieved from the model.
    \item Noisy gradients: Noise is added to the original gradients. The modified gradients are then used to update the model weights during training.
    \item Noisy labels: For noisy labels, the random value is added to the labels before training. We also included noise clipping to ensure the labels were within the correct range to avoid extreme changes and too much distortion in the labels. 
\end{itemize}

\begin{figure*}[ht]
\centering
\includegraphics[width = 1 \linewidth, keepaspectratio]{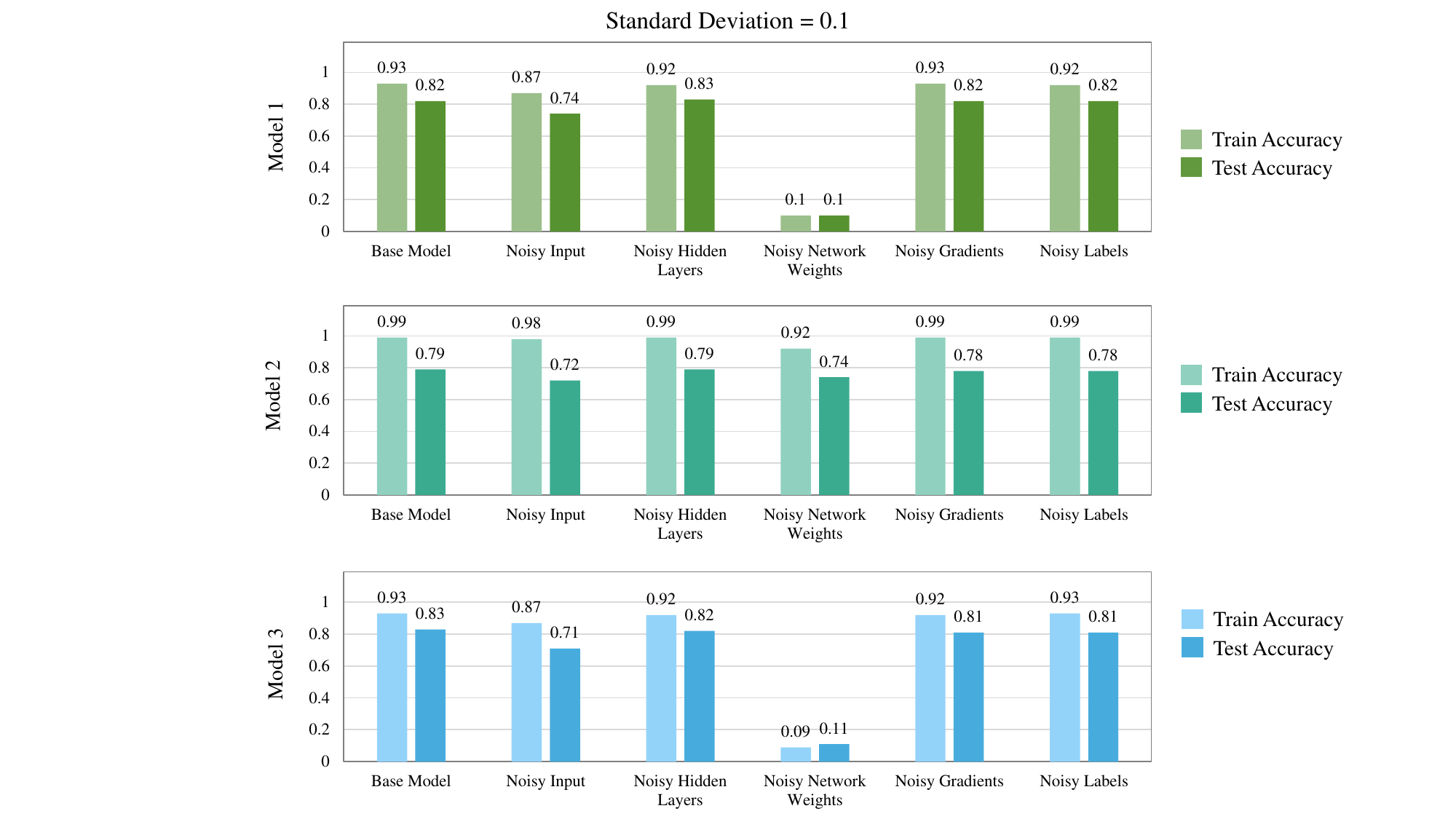}
\caption{This figure presents a comparison of the training and test accuracy of three models across six mechanisms. The first set of columns for each figure represents the base model trained without noise.}
\label{Model comparison-5}
\end{figure*}

In this section, We explore the effectiveness of different mechanisms and compare their results with those of Gaussian noise hidden layers. We train the centralized CNN models using five noise infusion mechanisms where the standard deviation of the additive noise is consistently set at $0.1$. The results are presented in Figure \ref{Model comparison-5}.

The base model serves as a control group without additive noise. Models 1 and 3 are most sensitive to injecting noise into input and weights, significantly dropping training and test accuracy. The models with noisy weights also fail to learn effectively and generalize, which indicates the detrimental impact of noisy weights on training. While there is a slight decrease in training and test accuracy, models trained with Gaussian noise hidden layers, labels, and gradients are less sensitive to noise. Model 2 is the most stable among the three, and the decrease in the accuracy is less significant. When the added noise's standard deviation is $0.1$, Gaussian noise hidden layers, noisy gradients, and noisy labels are the most resilient. Hence, we continue studying these models under varying noise levels. 

\begin{figure*}[ht]
\centering
\includegraphics[width = 1 \linewidth, keepaspectratio]{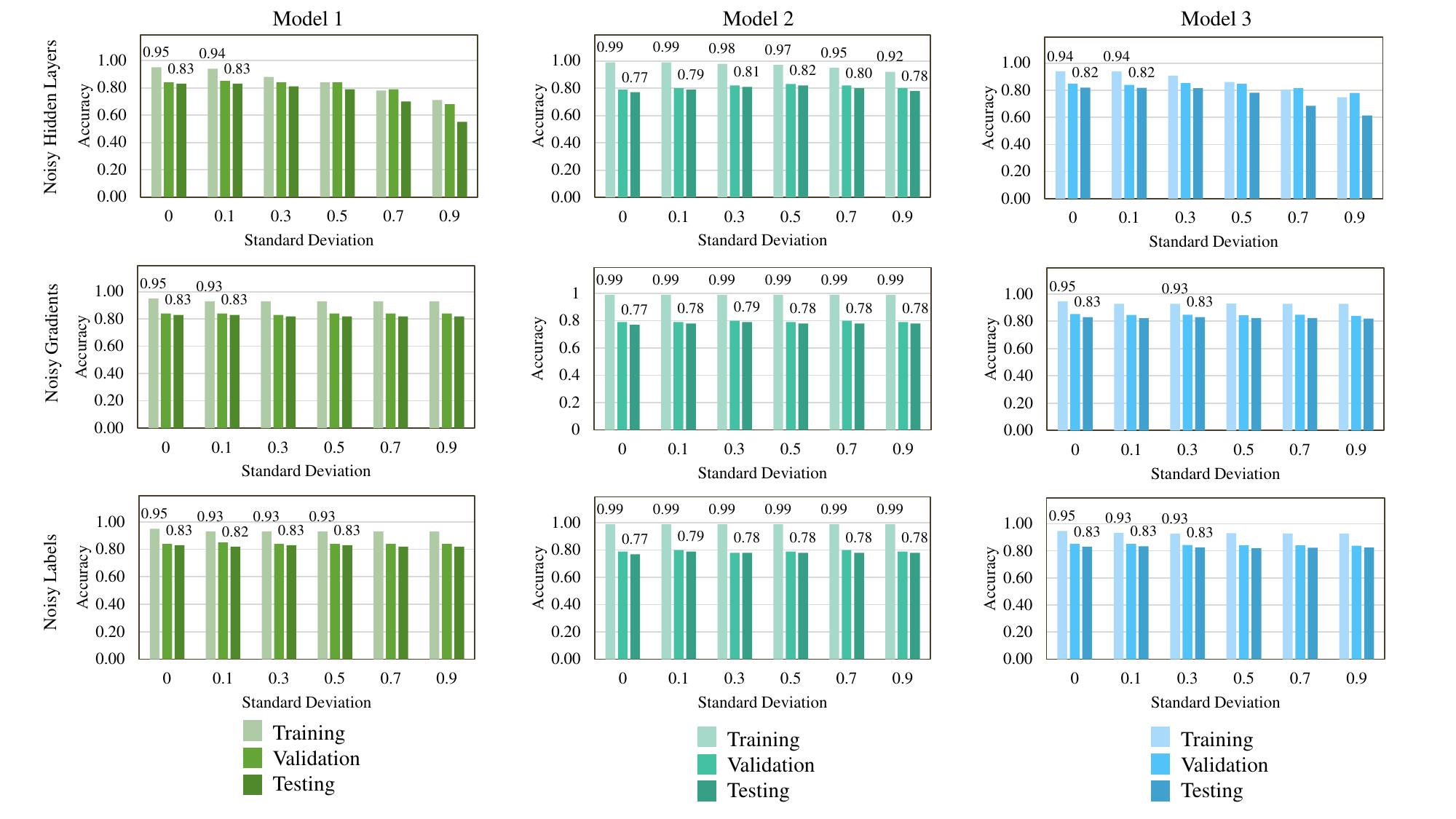}
\caption{Results of training and evaluating models with top 3 noise infusion methods and varying noise levels.}
\label{Model comparison-6}
\end{figure*}

The results from training the centralized data with Gaussian noise hidden layers, noisy gradients, and noisy labels using the three models are presented in Figure \ref{Model comparison-6}. The noise levels are $\sigma$ = $\{ 0.1, 0.3, 0.5, 0.7, 0.9\}$. 

While all models are somewhat sensitive to additive noise, they exhibit different performance variations under the noise infusion mechanisms. We can increase the noise in the models with Gaussian noise hidden layers while preserving test accuracy, especially in model 2, where the accuracy remains relatively constant compared to the base model. It is also interesting to see that increasing noise improves test accuracy, indicating the regularization effect of noise. In models 1 and 3, we can increase the noise levels to 0.1, and the test accuracy is 0.83 and 0.82, respectively.   
 
The models exhibit similar performance with noisy gradients. With models 1 and 3, the test accuracy gradually decreases as we increase the noise. However, even with a high standard deviation, model 2 remains stable against additive noise. 
Models 1 and 3, trained on data with noisy labels, have better stability, and we can increase the standard deviation to 0.5 and 0.3, respectively. Model 2 performs equally well when trained with noisy labels compared to the base model, demonstrating resilience to label noise.

Noise can negatively impact both training and test accuracy. However, the impact of noise on model performance varies depending on the noise infusion mechanism and the standard deviation of the additive noise. While the model's performance gradually degrades, noise can be used as a regularization technique. The results indicate that models with Gaussian noise hidden layers are effective in remaining stable even when the standard deviation of the noise is high.
\section{Conclusion}\label{sec-conclusion}
The present work is an empirical study on the role of noise in enhancing generalization, stability, and privacy within CNN for image classification. Through a series of carefully designed experiments, we observed that the introduction of noise during training helps prevent overfitting by making the model less reliant on precise features and patterns in the training data. By encouraging the network to learn more robust representations of data, CNNs with Gaussian noise hidden layers tend to perform better on unseen or noisy data, making them particularly useful for tasks where the input data may contain variations or uncertainties. This technique can improve CNN's ability to handle real-world scenarios and noisy environments, making it a valuable tool in applications such as image recognition, denoising, and signal processing. Our experimental results demonstrate that when the model is not over-parameterized, perturbing the parameters associated with the deep learning model by adding Gaussian noise behaves as an implicit regularization technique.  Models trained with noise generalize better, achieve higher accuracy, and are more stable in centralized and federated settings.

Introducing SNR as a measure of the signal quality (the base model performance) relative to the noise (noisy model performance) serves as a powerful tool for balancing accuracy and privacy in privacy-preserving settings. Additionally, PoS and PoA provide an in-depth understanding of the interplay of utility, stability, and privacy under different conditions. PoS and PoA can be used as tangible metrics for assessing the trade-off between privacy and accuracy in privacy-aware machine learning. 

Furthermore, we conducted a comparative analysis over CNN-based image classification noise infusion scenarios to determine the most effective methods of enhancing generalization, stability, and privacy. This investigation particularly benefits federated learning, where higher noise levels offer stronger privacy guarantees. 

This study has significant implications for practical machine learning applications that require reliable performance under varying conditions. Noise-infused models can help achieve models capable of handling diverse and noisy datasets. 

In the context of federated learning, understanding the impact of noise leads to designing computationally efficient private models.  The findings of this study demonstrate the potential of noise as a privacy-enhancing mechanism that can empower individuals and organizations to make informed decisions regarding data sharing and model deployment. By incorporating privacy-preserving techniques and acknowledging privacy as a fundamental human right, this research contributes to the responsible and ethical use of data and machine learning technologies.
\backmatter
\section*{Declarations}
\textbf{Conflict of interest} The authors declare no competing interests. \newline
\textbf{Consent to participate } Not applicable \newline
\textbf{Consent for publication} Not applicable\newline
\textbf{Ethics Approval} Not applicable\newline
\textbf{Funding} Not applicable \newline
\textbf{Data availability} The dataset analyzed during the current study is publicly available in the repository created by Alex Krizhevsky. \newline
\href{https://www.cs.toronto.edu/~kriz/cifar.html}{https://www.cs.toronto.edu/~kriz/cifar.html}\newline
\textbf{Code availability} The code is available in the author's GitHub repository.\newline 
\bibliography{bibliography}

\begin{thebibliography}{10}

\bibitem{bartlett2002rademacher}
Peter~L Bartlett and Shahar Mendelson.
\newblock Rademacher and gaussian complexities: Risk bounds and structural results.
\newblock {\em Journal of Machine Learning Research}, 3(Nov):463--482, 2002.

\bibitem{gnecco2008approximation}
Giorgio Gnecco, Marcello Sanguineti, et~al.
\newblock Approximation error bounds via rademacher complexity.
\newblock {\em Applied Mathematical Sciences}, 2:153--176, 2008.

\bibitem{ledoux1991probability}
Michel Ledoux and Michel Talagrand.
\newblock {\em Probability in Banach Spaces: isoperimetry and processes}, volume~23.
\newblock Springer Science \& Business Media, 1991.

\bibitem{mohri2008rademacher}
Mehryar Mohri and Afshin Rostamizadeh.
\newblock Rademacher complexity bounds for non-iid processes.
\newblock {\em Advances in Neural Information Processing Systems}, 21, 2008.

\bibitem{mohri2018foundations}
Mehryar Mohri, Afshin Rostamizadeh, and Ameet Talwalkar.
\newblock {\em Foundations of machine learning}.
\newblock MIT press, 2018.

\bibitem{vapnik2013uniform}
Vladimir~N Vapnik and Alexey~Ya Chervonenkis.
\newblock On the uniform convergence of the frequencies of occurrence of events to their probabilities.
\newblock In {\em Empirical Inference}, pages 7--12. Springer, 2013.

\bibitem{cybenko1996just}
George Cybenko.
\newblock Just-in-time learning and estimation.
\newblock {\em Nato ASI Series F Computer and Systems Sciences}, 153:423--434, 1996.

\bibitem{karpinski1995bounding}
Marek Karpinski and Angus Macintyre.
\newblock Bounding vc-dimension for neural networks: progress and prospects.
\newblock In {\em European Conference on Computational Learning Theory}, pages 337--341. Springer, 1995.

\bibitem{sontag1998vc}
Eduardo~D Sontag et~al.
\newblock Vc dimension of neural networks.
\newblock {\em NATO ASI Series F Computer and Systems Sciences}, 168:69--96, 1998.

\bibitem{goodfellow2016deep}
Ian Goodfellow, Yoshua Bengio, and Aaron Courville.
\newblock {\em Deep learning}.
\newblock MIT press, 2016.

\bibitem{rakhlin2005stability}
Alexander Rakhlin, Sayan Mukherjee, and Tomaso Poggio.
\newblock Stability results in learning theory.
\newblock {\em Analysis and Applications}, 3(04):397--417, 2005.

\bibitem{bousquet2002stability}
Olivier Bousquet and Andr{\'e} Elisseeff.
\newblock Stability and generalization.
\newblock {\em The Journal of Machine Learning Research}, 2:499--526, 2002.

\bibitem{bonnans1998optimization}
J~Fr{\'e}d{\'e}ric Bonnans and Alexander Shapiro.
\newblock Optimization problems with perturbations: A guided tour.
\newblock {\em SIAM review}, 40(2):228--264, 1998.

\bibitem{gavison1980privacy}
Ruth Gavison.
\newblock Privacy and the limits of law.
\newblock {\em The Yale law journal}, 89(3):421--471, 1980.

\bibitem{tene2011privacy}
Omer Tene and Jules Polonetsky.
\newblock Privacy in the age of big data: a time for big decisions.
\newblock {\em Stan. L. Rev. Online}, 64:63, 2011.

\bibitem{brandeis1890right}
Louis Brandeis and Samuel Warren.
\newblock The right to privacy.
\newblock {\em Harvard law review}, 4(5):193--220, 1890.

\bibitem{dinur2003revealing}
Irit Dinur and Kobbi Nissim.
\newblock Revealing information while preserving privacy.
\newblock In {\em Proceedings of the twenty-second ACM SIGMOD-SIGACT-SIGART symposium on Principles of database systems}, pages 202--210, 2003.

\bibitem{blum2005practical}
Avrim Blum, Cynthia Dwork, Frank McSherry, and Kobbi Nissim.
\newblock Practical privacy: the sulq framework.
\newblock In {\em Proceedings of the twenty-fourth ACM SIGMOD-SIGACT-SIGART symposium on Principles of database systems}, pages 128--138, 2005.

\bibitem{dwork2006calibrating}
Cynthia Dwork, Frank McSherry, Kobbi Nissim, and Adam Smith.
\newblock Calibrating noise to sensitivity in private data analysis.
\newblock In {\em Theory of cryptography conference}, pages 265--284. Springer, 2006.

\bibitem{dwork2011differential}
Cynthia Dwork, Frank McSherry, Kobbi Nissim, and Adam Smith.
\newblock Differential privacy—a primer for the perplexed,”.
\newblock {\em Joint UNECE/Eurostat work session on statistical data confidentiality}, 11, 2011.

\bibitem{dwork2014algorithmic}
Cynthia Dwork, Aaron Roth, et~al.
\newblock The algorithmic foundations of differential privacy.
\newblock {\em Foundations and Trends{\textregistered} in Theoretical Computer Science}, 9(3--4):211--407, 2014.

\bibitem{wei2020federated}
Kang Wei, Jun Li, Ming Ding, Chuan Ma, Howard~H Yang, Farhad Farokhi, Shi Jin, Tony~QS Quek, and H~Vincent Poor.
\newblock Federated learning with differential privacy: Algorithms and performance analysis.
\newblock {\em IEEE Transactions on Information Forensics and Security}, 15:3454--3469, 2020.

\bibitem{dwork2006our}
Cynthia Dwork, Krishnaram Kenthapadi, Frank McSherry, Ilya Mironov, and Moni Naor.
\newblock Our data, ourselves: Privacy via distributed noise generation.
\newblock In {\em Annual international conference on the theory and applications of cryptographic techniques}, pages 486--503. Springer, 2006.

\bibitem{mcmahan2017communication}
Brendan McMahan, Eider Moore, Daniel Ramage, Seth Hampson, and Blaise~Aguera y~Arcas.
\newblock Communication-efficient learning of deep networks from decentralized data.
\newblock In {\em Artificial intelligence and statistics}, pages 1273--1282. PMLR, 2017.

\bibitem{bonawitz2019towards}
Keith Bonawitz, Hubert Eichner, Wolfgang Grieskamp, Dzmitry Huba, Alex Ingerman, Vladimir Ivanov, Chloe Kiddon, Jakub Kone{\v{c}}n{\`y}, Stefano Mazzocchi, Brendan McMahan, et~al.
\newblock Towards federated learning at scale: System design.
\newblock {\em Proceedings of machine learning and systems}, 1:374--388, 2019.

\bibitem{kairouz2021advances}
Peter Kairouz, H~Brendan McMahan, Brendan Avent, Aur{\'e}lien Bellet, Mehdi Bennis, Arjun~Nitin Bhagoji, Kallista Bonawitz, Zachary Charles, Graham Cormode, Rachel Cummings, et~al.
\newblock Advances and open problems in federated learning.
\newblock {\em Foundations and Trends{\textregistered} in Machine Learning}, 14(1--2):1--210, 2021.

\bibitem{ron1999algorithmic}
Dana Ron and M~Kearns.
\newblock Algorithmic stability and sanity-check bounds for leave-one-out crossvaildation.
\newblock {\em Neural Computation}, 11(6):1427--1453, 1999.

\bibitem{neyshabur2017exploring}
Behnam Neyshabur, Srinadh Bhojanapalli, David McAllester, and Nati Srebro.
\newblock Exploring generalization in deep learning.
\newblock {\em Advances in neural information processing systems}, 30, 2017.

\bibitem{bishop1995training}
Chris~M Bishop.
\newblock Training with noise is equivalent to tikhonov regularization.
\newblock {\em Neural computation}, 7(1):108--116, 1995.

\bibitem{bishop1995neural}
Christopher~M Bishop et~al.
\newblock {\em Neural networks for pattern recognition}.
\newblock Oxford University Press, 1995.

\bibitem{shalev2014understanding}
Shai Shalev-Shwartz and Shai Ben-David.
\newblock {\em Understanding machine learning: From theory to algorithms}.
\newblock Cambridge University Press, 2014.

\bibitem{mcdonnell2011benefits}
Mark~D McDonnell and Lawrence~M Ward.
\newblock The benefits of noise in neural systems: bridging theory and experiment.
\newblock {\em Nature Reviews Neuroscience}, 12(7):415--425, 2011.

\bibitem{doyle2018colour}
J~Andrew Doyle and Alan~C Evans.
\newblock What colour is neural noise?
\newblock {\em arXiv preprint arXiv:1806.03704}, 2018.

\bibitem{kumar2019novel}
Sumit Kumar, Ayush Kumar, and Rajib~Kumar Jha.
\newblock A novel noise-enhanced back-propagation technique for weak signal detection in neyman--pearson framework.
\newblock {\em Neural Processing Letters}, 50(3):2389--2406, 2019.

\bibitem{benzi1981mechanism}
Roberto Benzi, Alfonso Sutera, and Angelo Vulpiani.
\newblock The mechanism of stochastic resonance.
\newblock {\em Journal of Physics A: mathematical and general}, 14(11):L453, 1981.

\bibitem{benzi1983theory}
Roberto Benzi, Giorgio Parisi, Alfonso Sutera, and Angelo Vulpiani.
\newblock A theory of stochastic resonance in climatic change.
\newblock {\em SIAM Journal on applied mathematics}, 43(3):565--578, 1983.

\bibitem{ikemoto2018noise}
Shuhei Ikemoto, Fabio DallaLibera, and Koh Hosoda.
\newblock Noise-modulated neural networks as an application of stochastic resonance.
\newblock {\em Neurocomputing}, 277:29--37, 2018.

\bibitem{faisal2008noise}
A~Aldo Faisal, Luc~PJ Selen, and Daniel~M Wolpert.
\newblock Noise in the nervous system.
\newblock {\em Nature reviews neuroscience}, 9(4):292--303, 2008.

\bibitem{maass2014noise}
Wolfgang Maass.
\newblock Noise as a resource for computation and learning in networks of spiking neurons.
\newblock {\em Proceedings of the IEEE}, 102(5):860--880, 2014.

\bibitem{holmstrom1992using}
Lasse Holmstrom and Petri Koistinen.
\newblock Using additive noise in back-propagation training.
\newblock {\em IEEE Transactions on Neural Networks}, 3(1):24--38, 1992.

\bibitem{karpukhin2019training}
Vladimir Karpukhin, Omer Levy, Jacob Eisenstein, and Marjan Ghazvininejad.
\newblock Training on synthetic noise improves robustness to natural noise in machine translation.
\newblock {\em arXiv preprint arXiv:1902.01509}, 2019.

\bibitem{sietsma1991creating}
Jocelyn Sietsma and Robert~JF Dow.
\newblock Creating artificial neural networks that generalize.
\newblock {\em Neural networks}, 4(1):67--79, 1991.

\bibitem{zeng2021noise}
Zhi Zeng, Yuan Liu, Weijun Tang, and Fangjiong Chen.
\newblock Noise is useful: Exploiting data diversity for edge intelligence.
\newblock {\em IEEE Wireless Communications Letters}, 10(5):957--961, 2021.

\bibitem{reed1999neural}
Russell Reed and Robert~J MarksII.
\newblock {\em Neural smithing: supervised learning in feedforward artificial neural networks}.
\newblock Mit Press, 1999.

\bibitem{zur2009noise}
Richard~M Zur, Yulei Jiang, Lorenzo~L Pesce, and Karen Drukker.
\newblock Noise injection for training artificial neural networks: A comparison with weight decay and early stopping.
\newblock {\em Medical physics}, 36(10):4810--4818, 2009.

\bibitem{nagabushan2016effect}
Naresh Nagabushan, Nishank Satish, and S~Raghuram.
\newblock Effect of injected noise in deep neural networks.
\newblock In {\em 2016 IEEE International Conference on Computational Intelligence and Computing Research (ICCIC)}, pages 1--5. IEEE, 2016.

\bibitem{dhifallah2021inherent}
Oussama Dhifallah and Yue Lu.
\newblock On the inherent regularization effects of noise injection during training.
\newblock In {\em International Conference on Machine Learning}, pages 2665--2675. PMLR, 2021.

\bibitem{he2019parametric}
Zhezhi He, Adnan~Siraj Rakin, and Deliang Fan.
\newblock Parametric noise injection: Trainable randomness to improve deep neural network robustness against adversarial attack.
\newblock In {\em Proceedings of the IEEE/CVF Conference on Computer Vision and Pattern Recognition}, pages 588--597, 2019.

\bibitem{xiao2021noise}
Li~Xiao, Zeliang Zhang, and Yijie Peng.
\newblock Noise optimization for artificial neural networks.
\newblock {\em arXiv preprint arXiv:2102.04450}, 2021.

\bibitem{liu2021training}
Aishan Liu, Xianglong Liu, Hang Yu, Chongzhi Zhang, Qiang Liu, and Dacheng Tao.
\newblock Training robust deep neural networks via adversarial noise propagation.
\newblock {\em IEEE Transactions on Image Processing}, 30:5769--5781, 2021.

\bibitem{koistinen1991}
P.~Koistinen and L.~Holmstrom.
\newblock Kernel regression and backpropagation training with noise.
\newblock In {\em [Proceedings] 1991 IEEE International Joint Conference on Neural Networks}, pages 367--372 vol.1, 1991.

\bibitem{matsuoka1992noise}
Kiyotoshi Matsuoka.
\newblock Noise injection into inputs in back-propagation learning.
\newblock {\em IEEE Transactions on Systems, Man, and Cybernetics}, 22(3):436--440, 1992.

\bibitem{abadi2016deep}
Martin Abadi, Andy Chu, Ian Goodfellow, H~Brendan McMahan, Ilya Mironov, Kunal Talwar, and Li~Zhang.
\newblock Deep learning with differential privacy.
\newblock In {\em Proceedings of the 2016 ACM SIGSAC conference on computer and communications security}, pages 308--318, 2016.

\bibitem{an1996effects}
Guozhong An.
\newblock The effects of adding noise during backpropagation training on a generalization performance.
\newblock {\em Neural computation}, 8(3):643--674, 1996.

\bibitem{wang1999training}
Chuan Wang and Jose~C Principe.
\newblock Training neural networks with additive noise in the desired signal.
\newblock {\em IEEE Transactions on Neural Networks}, 10(6):1511--1517, 1999.

\bibitem{azamimi2010effect}
Azian Azamimi, Yoko Uwate, and Yoshifumi Nishio.
\newblock Effect of chaos noise on the learning ability of back propagation algorithm in feed forward neural network.
\newblock In {\em 2010 6th International Colloquium on Signal Processing \& its Applications}, pages 1--4. IEEE, 2010.

\bibitem{alonso2014combining}
Juan~Manuel Alonso-Weber, MP~Sesmero, and Araceli Sanchis.
\newblock Combining additive input noise annealing and pattern transformations for improved handwritten character recognition.
\newblock {\em Expert systems with applications}, 41(18):8180--8188, 2014.

\bibitem{isaev2016training}
IV~Isaev and SA~Dolenko.
\newblock Training with noise as a method to increase noise resilience of neural network solution of inverse problems.
\newblock {\em Optical Memory and Neural Networks}, 25(3):142--148, 2016.

\bibitem{kosko2020noise}
Bart Kosko, Kartik Audhkhasi, and Osonde Osoba.
\newblock Noise can speed backpropagation learning and deep bidirectional pretraining.
\newblock {\em Neural Networks}, 129:359--384, 2020.

\bibitem{brown2003use}
Warick~M Brown, Tam{\'a}s~D Gedeon, and David~I Groves.
\newblock Use of noise to augment training data: a neural network method of mineral--potential mapping in regions of limited known deposit examples.
\newblock {\em Natural Resources Research}, 12(2):141--152, 2003.

\bibitem{hua2006noise}
Jianping Hua, James Lowey, Zixiang Xiong, and Edward~R Dougherty.
\newblock Noise-injected neural networks show promise for use on small-sample expression data.
\newblock {\em BMC bioinformatics}, 7(1):1--14, 2006.

\bibitem{li2016whiteout}
Yinan Li and Fang Liu.
\newblock Whiteout: Gaussian adaptive noise regularization in deep neural networks.
\newblock {\em arXiv preprint arXiv:1612.01490}, 2016.

\bibitem{lim2021noisy}
Soon~Hoe Lim, N~Benjamin Erichson, Francisco Utrera, Winnie Xu, and Michael~W Mahoney.
\newblock Noisy feature mixup.
\newblock {\em arXiv preprint arXiv:2110.02180}, 2021.

\bibitem{arani2021noise}
Elahe Arani, Fahad Sarfraz, and Bahram Zonooz.
\newblock Noise as a resource for learning in knowledge distillation.
\newblock In {\em Proceedings of the IEEE/CVF Winter Conference on Applications of Computer Vision}, pages 3129--3138, 2021.

\bibitem{you2019adversarial}
Zhonghui You, Jinmian Ye, Kunming Li, Zenglin Xu, and Ping Wang.
\newblock Adversarial noise layer: Regularize neural network by adding noise.
\newblock In {\em 2019 IEEE International Conference on Image Processing (ICIP)}, pages 909--913. IEEE, 2019.

\bibitem{adilova2018introducing}
Linara Adilova, Nathalie Paul, and Peter Schlicht.
\newblock Introducing noise in decentralized training of neural networks.
\newblock In {\em Joint European Conference on Machine Learning and Knowledge Discovery in Databases}, pages 37--48. Springer, 2018.

\bibitem{sapkal2018modified}
Ashwini Sapkal and UV~Kulkarni.
\newblock Modified backpropagation with added white gaussian noise in weighted sum for convergence improvement.
\newblock {\em Procedia computer science}, 143:309--316, 2018.

\bibitem{shi2020anti}
Jiashuo Shi, Mingce Chen, Dong Wei, Chai Hu, Jun Luo, Haiwei Wang, Xinyu Zhang, and Changsheng Xie.
\newblock Anti-noise diffractive neural network for constructing an intelligent imaging detector array.
\newblock {\em Optics Express}, 28(25):37686--37699, 2020.

\bibitem{bykov2021noisegrad}
Kirill Bykov, Anna Hedstr{\"o}m, Shinichi Nakajima, and Marina M-C H{\"o}hne.
\newblock Noisegrad: enhancing explanations by introducing stochasticity to model weights.
\newblock {\em arXiv preprint arXiv:2106.10185}, 2021.

\bibitem{edwards1998fault}
Peter~J Edwards and Alan~F Murray.
\newblock Fault tolerance via weight noise in analog vlsi implementations of mlps-a case study with epsilon.
\newblock {\em IEEE Transactions on Circuits and Systems II: Analog and Digital Signal Processing}, 45(9):1255--1262, 1998.

\bibitem{zhou2019toward}
Mo~Zhou, Tianyi Liu, Yan Li, Dachao Lin, Enlu Zhou, and Tuo Zhao.
\newblock Toward understanding the importance of noise in training neural networks.
\newblock In {\em International Conference on Machine Learning}, pages 7594--7602. PMLR, 2019.

\bibitem{duan2021noise}
Lingling Duan, Fabing Duan, Fran{\c{c}}ois Chapeau-Blondeau, and Derek Abbott.
\newblock Noise-boosted backpropagation learning of feedforward threshold neural networks for function approximation.
\newblock {\em IEEE Transactions on Instrumentation and Measurement}, 70:1--12, 2021.

\bibitem{chaudhari2015effect}
PRATIK Chaudhari and STEFANO Soatto.
\newblock The effect of gradient noise on the energy landscape of deep networks.
\newblock Technical report, Technical Report Preprint, 2015.

\bibitem{neelakantan2015adding}
Arvind Neelakantan, Luke Vilnis, Quoc~V Le, Ilya Sutskever, Lukasz Kaiser, Karol Kurach, and James Martens.
\newblock Adding gradient noise improves learning for very deep networks.
\newblock {\em arXiv preprint arXiv:1511.06807}, 2015.

\bibitem{mirzasoleiman2020coresets}
Baharan Mirzasoleiman, Kaidi Cao, and Jure Leskovec.
\newblock Coresets for robust training of deep neural networks against noisy labels.
\newblock {\em Advances in Neural Information Processing Systems}, 33:11465--11477, 2020.

\bibitem{jiang2022towards}
Xuefeng Jiang, Sheng Sun, Yuwei Wang, and Min Liu.
\newblock Towards federated learning against noisy labels via local self-regularization.
\newblock In {\em Proceedings of the 31st ACM International Conference on Information \& Knowledge Management}, pages 862--873, 2022.

\bibitem{wu2022stgn}
Tingting Wu, Xiao Ding, Minji Tang, Hao Zhang, Bing Qin, and Ting Liu.
\newblock Stgn: an implicit regularization method for learning with noisy labels in natural language processing.
\newblock In {\em Proceedings of the 2022 Conference on Empirical Methods in Natural Language Processing}, pages 7587--7598, 2022.

\bibitem{song2022learning}
Hwanjun Song, Minseok Kim, Dongmin Park, Yooju Shin, and Jae-Gil Lee.
\newblock Learning from noisy labels with deep neural networks: A survey.
\newblock {\em IEEE Transactions on Neural Networks and Learning Systems}, 2022.

\bibitem{smith1997scientist}
Steven~W Smith et~al.
\newblock The scientist and engineer's guide to digital signal processing, 1997.

\bibitem{mazda2014telecommunications}
Fraidoon Mazda.
\newblock {\em Telecommunications engineer's reference book}.
\newblock Butterworth-Heinemann, 2014.

\bibitem{anshelevich2008price}
Elliot Anshelevich, Anirban Dasgupta, Jon Kleinberg, {\'E}va Tardos, Tom Wexler, and Tim Roughgarden.
\newblock The price of stability for network design with fair cost allocation.
\newblock {\em SIAM Journal on Computing}, 38(4):1602--1623, 2008.

\bibitem{koutsoupias2009worst}
Elias Koutsoupias and Christos Papadimitriou.
\newblock Worst-case equilibria.
\newblock {\em Computer science review}, 3(2):65--69, 2009.

\bibitem{zhang2021understanding}
Chiyuan Zhang, Samy Bengio, Moritz Hardt, Benjamin Recht, and Oriol Vinyals.
\newblock Understanding deep learning (still) requires rethinking generalization.
\newblock {\em Communications of the ACM}, 64(3):107--115, 2021.

\bibitem{simonyan2014very}
Karen Simonyan and Andrew Zisserman.
\newblock Very deep convolutional networks for large-scale image recognition.
\newblock {\em arXiv preprint arXiv:1409.1556}, 2014.

\end{thebibliography}
\end{document}